\documentclass[letterpaper]{article} 
\usepackage{aaai23}  
\usepackage{times}  
\usepackage{helvet}  
\usepackage{courier}  
\usepackage[hyphens]{url}  
\usepackage{graphicx} 
\usepackage{natbib}  
\usepackage{caption} 
\usepackage{algorithm}
\usepackage{algorithmic}

\usepackage{newfloat}
\usepackage{listings}
\DeclareCaptionStyle{ruled}{labelfont=normalfont,labelsep=colon,strut=off} 
\floatstyle{ruled}
\newfloat{listing}{tb}{lst}{}
\floatname{listing}{Listing} 
\nocopyright 

\usepackage{amsmath,amssymb,amsfonts}
\usepackage{booktabs}
\usepackage{nicefrac}
\usepackage{microtype}
\usepackage{textcomp}
\usepackage{todonotes}
\usepackage{subcaption}
\usepackage{svg}
\usepackage{enumitem}
\usepackage{multirow}
\usepackage[hidelinks]{hyperref}
\usepackage{url}

\title{Towards Hardware-Specific Automatic Compression of Neural Networks}
\author{
    Torben Krieger\equalcontrib,
    Bernhard Klein\equalcontrib\href{https://orcid.org/0000-0003-0497-5748}{\includegraphics[scale=0.08]{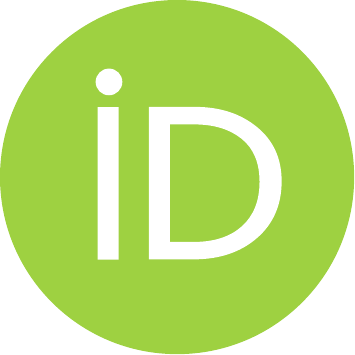}},
    Holger Fröning\hspace{0.5mm}\href{https://orcid.org/0000-0001-9562-0680}{\includegraphics[scale=0.08]{orcid.pdf}}
}
\affiliations{
    Institute of Computer Engineering\\
    Heidelberg University, Germany\\
    torben.krieger@stud.uni-heidelberg.de, bernhard.klein@ziti.uni-heidelberg.de, holger.froening@ziti.uni-heidelberg.de

}

\begin{document}

\maketitle

\begin{abstract}
Compressing neural network architectures is important to allow the deployment of models to embedded or mobile devices, 
and pruning and quantization are the major approaches to compress neural networks nowadays.
Both methods benefit when compression parameters are selected specifically for each layer.
Finding good combinations of compression parameters, so-called compression policies, is hard as the problem spans an exponentially large search space.
Effective compression policies consider the influence of the specific hardware architecture on the used compression methods.
We propose an algorithmic framework called "Galen"\footnote{Publicly available: \url{https://github.com/UniHD-CEG/galen}}\footnote{Galen (129 C.E. – c. 216 C.E.) was a Greek physician and philosopher who is regarded as a pioneer in surgery. He valued observation, experimentation and analysis to advance his studies. Similar to Galen, the present work observes sensitivity to guide compression, and includes experimentation to assess hardware costs before re-assessing the strategy of network alteration (surgery).}
to search such policies using reinforcement learning utilizing pruning and quantization, thus providing automatic compression for neural networks.
Contrary to other approaches we use inference latency measured on the target hardware device as an optimization goal.
With that, the framework supports the compression of models specific to a given hardware target.
We validate our approach using three different reinforcement learning agents for pruning, quantization and joint pruning and quantization.
Besides proving the functionality of our approach we were able to compress a ResNet18 for CIFAR-10, on an embedded ARM processor, to 20\% of the original inference latency without significant loss of accuracy.
Moreover, we can demonstrate that a joint search and compression using pruning and quantization is superior to an individual search for policies using a single compression method.

\end{abstract}

\section{Introduction}
While the success of machine learning press deep neural networks forward to various problem domains, the deployment on resource-constrained embedded or mobile devices is limited due to its high compute demands.
This contradicts the practical application of deep learning approaches to real-world problems, as inference with such models does not provide acceptable latencies, or is too costly in terms of energy demand for battery-powered devices. 
Well-known compression methods like pruning or quantization can improve the hardware efficiency significantly~\cite{heChannelPruningAccelerating2017,jacobQuantizationTrainingNeural2018}.
However, applying the same compression parameters---specifying sparsity for pruning and precision for quantization---to all layers of a network yields suboptimal results, since the computational complexity and sensitivity differs highly between layers. 
Therefore, a mixed compression policy specifying layer-specific compression parameters is required to achieve near-optimal compression results while maintaining top accuracy.
Searching compression parameters per layer spans exponentially large search spaces prohibiting the use of structured search methods.
Applying pruning and quantization at the same time makes the problem even more complex due to reciprocal effects.
The search by a human expert applying heuristics gained from experience mismatches the scalability demands and is insufficient due to the limited availability of such experts.

Various approaches were proposed to find compression policies automatically for either pruning or quantization.
Some select layer policies in a greedy fashion to fulfill a constraint \cite{yangNetAdaptPlatformawareNeural2018},
while others propose differential solutions by adding additional losses \cite{yuACPAutomaticChannel2022}.
Specifically for quantization, there are multiple approaches for selecting proper compression parameters by computing layer sensitivity metrics \cite{caiZeroQNovelZero2020,dongHAWQHessianAWare2019,dongHAWQV2HessianAware2020}.
Also heuristic search or evolutionary algorithms were prosed to find optimal solutions \cite{liuAutoCompressAutomaticDNN2020,linChannelPruningAutomatic2020}.
Besides searching for a separate policy per method some approaches are searching jointly for a combined compression policy \cite{yangAutomaticNeuralNetwork2020,tungCLIPQDeepNetwork2018, wangDifferentiableJointPruning2020, wangAPQJointSearch2020}.
As we will see, particularly interesting in the context of this work are approaches using reinforcement learning to predict compression policies \cite{wangHAQHardwareAwareAutomated2019,heAMCAutoMLModel2019,louAutoQAutomatedKernelwise2020,elthakebReLeQReinforcementLearning2020},
with in particular AMC~\cite{heAMCAutoMLModel2019} for pruning and HAQ~\cite{wangHAQHardwareAwareAutomated2019} for quantization demonstrating promising results.
Both process the models in a layer-wise fashion and predict the compression parameters as continuous actions using a reinforcement agent implementing the Deep Deterministic Policy Gradient (DDPG)
algorithm~\cite{lillicrapContinuousControlDeep2019}.

While reinforcement learning has demonstrated promising results for either searching pruning or quantization policies, there is no prior work yet on joint searches based on reinforcement learning.
A joint search is essential to cover reciprocal effects of applying quantization and pruning to the same model.
The introduction of a huge pruning sparsity might, for example, prohibit quantization to a layer weight, although strong quantization could yield better accuracy.
A joint search approach could consider such effects, while executing different policy searches separately might miss it.

Additionally, the reward for reinforcement learning is usually based on rather abstract metrics, instead of probing an actual targeted hardware instance for feedback in the form of latency, for instance.
Most related works use abstract metrics such as MACs~(Multiply-Accumulate Operations) or BOPs~(Bit OPerations)~\cite{heAMCAutoMLModel2019,wangDifferentiableJointPruning2020}, however, common abstract metrics do not directly translate to latency. 
Often the specific hardware architecture, through complex interaction of caches, memory bandwidths, etc. interacting with the parallel execution model, leads to a non-trivial correlation between metric and latency~\cite{Klein2021CacheBoundness,sze2020dlprocessormetric}.
In some cases a metric like BOPs, indicate a high speedup, while the used hardware did not support quantized data types or the overhead of using quantization methods---like bit-serial approaches~\cite{umurogluOptimizingBitSerialMatrix2019}---is much larger then the benefit.
Other works use lookup tables with latencies measured for specific layer configurations upfront~\cite{wangHAQHardwareAwareAutomated2019,yangNetAdaptPlatformawareNeural2018,wangAPQJointSearch2020}.
By definition, these lookup tables can only hold results for different configurations of individual layers.
Still, the size and effort for creating such a lookup table for a joint search problem is impractical due to the increased number of options per layer.
In addition, we can imagine that lookup tables could fail estimating latencies properly due to effects of layer combinations.

With this work, we propose an algorithm called "Galen" for a joint search of quantization and pruning policies, which also considers a reward based on actual target hardware latency.
The algorithm consists of a reinforcement learning agent to predict policies, which will be tested on a target device to integrate latency as a cost factor into the reward function.
Therefore, Galen finds hardware-specific policies and enables fast deployment to specific hardware devices.
We show exemplary its applicability to arbitrary trained image classification models to automatically search compression policies using pruning and quantization.
In particular, this work makes the following contributions:

\begin{enumerate}
    \item A generic algorithm ("Galen") and implementation for joint pruning and quantization based on reinforcement learning, supporting arbitrary models for image classification.
    \item Integration of direct hardware feedback by measuring model architecture latency on the target device, using hardware-specific code generation with support for sparse and quantized models.
    \item Proposals for three different reinforcement agents for pruning, quantization, and joint pruning and quantization.
    \item Support for quantization based on integer 8-bit and flexible bit widths, applicable interchangeably within a model, and conceptionally extendible to other data types.
\end{enumerate}
Our algorithm has its conceptual foundation in the ideas proposed by AMC~\cite{heAMCAutoMLModel2019} and HAQ~\cite{wangHAQHardwareAwareAutomated2019}.
This work will elaborate on the conceptual construction of the algorithm and the different agents proposed.
Within the evaluation, we will show why a joint hardware-specific approach using reinforcement learning is valuable.

\section{Related Work}
Denoted as \textit{AutoML} or \textit{Automatic Compression} various approaches for searching pruning or quantization policies automatically were proposed,
which mainly differ in solving the underlying optimization problem.
Related work covers relatively simple greedy algorithms (\textit{NetAdapt}~\cite{yangNetAdaptPlatformawareNeural2018}), reinforcement learning (\textit{AMC}~\cite{heAMCAutoMLModel2019}), simulated annealing (\textit{AutoCompress}~\cite{liuAutoCompressAutomaticDNN2020}), evolutionary algorithms (\textit{Automatic Structure Search}~\cite{linChannelPruningAutomatic2020}), among others. 
Besides NetAdapt all presented algorithms use an indirect metric as cost measurement within the optimization problem, e.g. the number of MACs, FLOPs or parameters.
Most approaches to search quantization policies with a specific precision per layer are either based on a metric measuring the sensitivity of a layer for quantization~\cite{caiZeroQNovelZero2020,dongHAWQHessianAWare2019,dongHAWQV2HessianAware2020} or use a reinforcement learning agent to predict policies~\cite{louAutoQAutomatedKernelwise2020,wangHAQHardwareAwareAutomated2019,elthakebReLeQReinforcementLearning2020}.
Also, it has been shown that the combination of both, pruning and quantization is very effective to achieve high compression ratios with low accuracy loss~\cite{hanDeepCompressionCompressing2016}.
In this regard, joint search has been considered in the form of Bayesian optimization (\textit{CLIP-Q}~\cite{tungCLIPQDeepNetwork2018}), gradient optimization~\cite{wangDifferentiableJointPruning2020}, and constrained optimization~\cite{yangAutomaticNeuralNetwork2020}.
Some of these approaches report results based on BOPs (Bit OPerations~\cite{Baskin2021bops}), but notably none of them measure the resulting latency or speedup.

In more detail, AMC~\cite{heAMCAutoMLModel2019} supports policies for structured pruning, based on input channels, or unstructured pruning by removing individual connections. However, the latter lacks speedup on real hardware and is less relevant in the context of this work.
HAQ~\cite{wangHAQHardwareAwareAutomated2019} produces mixed-precision policies covering bit widths from 2 to 8 bits for weights and activations of supported layers, thus always compresses to at least 8 bits.
Both algorithms follow the same schema and process a model layer-by-layer.
A layer-specific state is constructed and passed to a reinforcement agent which predicts the compression action for the layer.
The actions of both approaches are continuous, and a DDPG agent consisting of an actor and a critic network is used.
Subsequently, the actions are mapped to discrete compression parameters, precisely channel count or bit width.
After parsing all layers, the complete policy is validated by compressing the model and testing the achieved accuracy.
While HAQ conducts a short retraining before validating the performance, AMC instead reconstructs weight values by using stored input and output data of each layer.

AMC and HAQ mainly use the validation error as a reward, thus the agent is penalized based on the loss of accuracy introduced by a predicted policy. 
As this provides no incentive to compress the model,
both approaches ensure compression by enforcing a \emph{hard cost constraint}, defined as a ratio of the cost metric: while AMC uses the number of FLOPs or parameters as a cost metric, HAQ uses the inference latency estimated by using a lookup table.

\section{Algorithmic Concept: A Generic Reinforcement Learning Compression Framework}
This work proposes a general method, which automatically predicts a policy of Compression Method Parameters (CMPs) leading to a near optimal solution, balancing accuracy and latency.
In this work the compression methods are applied layer-wise, therefore, we define the compression policy ${P}$ compressing a model $\mathcal{M}$ to  $\mathcal{M}_{{P}}$ as,
 \begin{equation}
    {P} \in \{ \mathbf{r} \in \mathbb{R}^K | r_i \in [0, 1]\}^{L \times M},
\end{equation}
where $L$ is the number of layers of the model, $M$ is the number of used compression methods and $\mathbf{r}$ is a vector of $K$ continuous compression parameters.
While most of the CMPs---like amount of pruned channels or bit width of weights or activations---are discrete values, the policy uses normalized, continuous values.
While evaluating a policy by applying a compression the continuous policy $P$ is mapped to the hardware and implementation-dependent CMPs.
This allows a unbiased policy search, independent of the magnitude and granularity of the parameters.
However, some methods require a individual decision, e.g. to use a specific data type, for those, we weaken the definition and use a dictionary type holding status flags per method to represent the policy.

With that, searching for the best compression policy $\hat{{P}}$ that fulfills a target compression rate $c$ could be formulated as a constrained optimization problem
\begin{equation}
\begin{split}
    \hat{P} = \ \underset{P}{\arg\max} \ \textit{acc}\left(\mathcal{M}_{{P}}\left(\theta; x\right), y\right),& \\
   s.t.\ \textit{cost}\left(\mathcal{M}_{{P}}\right) \leq c \cdot \textit{cost}\left(\mathcal{M}\right),&
\end{split}
\end{equation}
where the output predicted by the compressed model $\mathcal{M}_{{P}}\left(\theta; x\right)$ for input $x$ and trained weights $\theta$, is validated with ground truth labels $y$ computing the accuracy $acc\left(\cdot\right)$ constrained by the selectable $cost\left(\cdot\right)$ metric. 

As cost metric, we use the inference latency of the compressed model.
The following presents the conceptual basics of our algorithm which utilizes reinforcement learning agents to predict a policy ${P}$.

\subsection{Algorithmic Schema}
\begin{figure}[h]
       \centering
      	\includegraphics[width=0.3\textwidth]{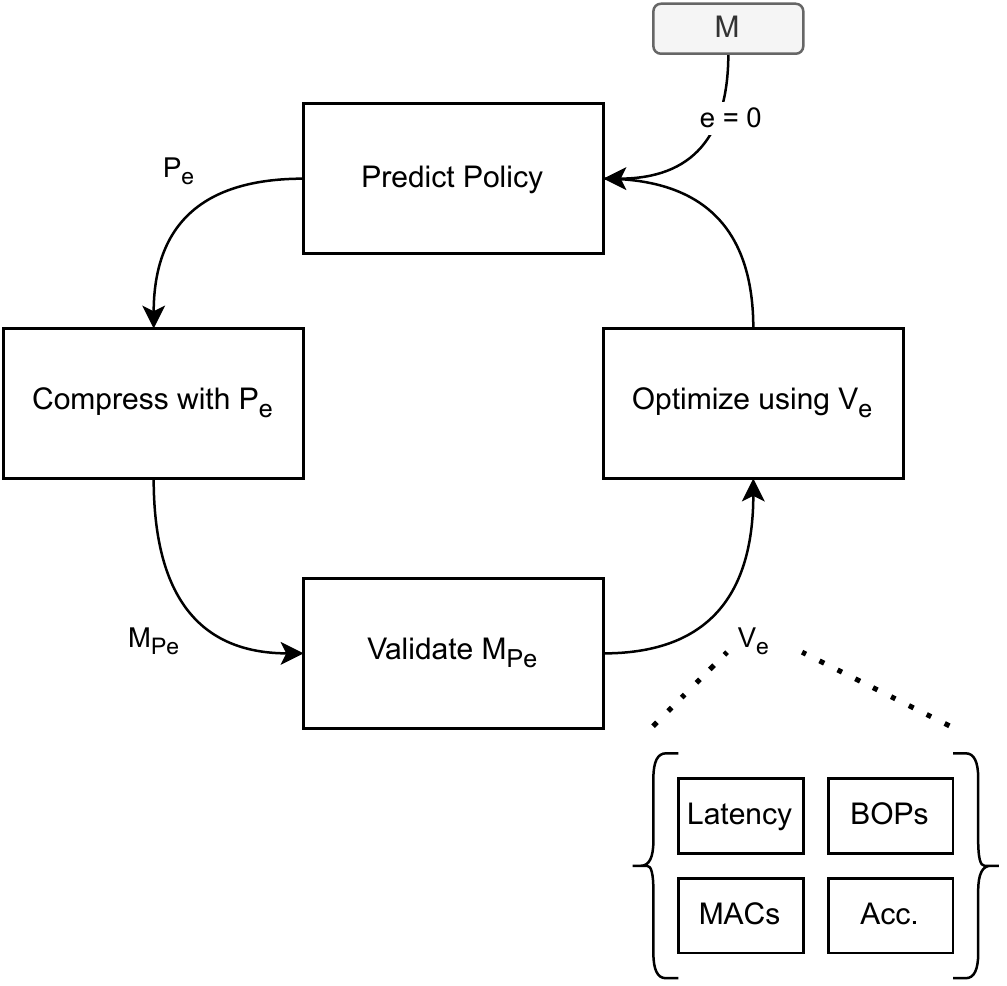}
	   \caption{Episode overview: predict, apply and validate a compression policy $P_e$ iteratively to optimize the agent.} 
	   \label{fig:concept:alg-schema:episode}       
\end{figure}

We distinguish between \textit{episodes}, the outer loop with hardware evaluation, and \textit{time steps}, the inner loop predicting policies for all layers.
Considering a reinforcement learning setup, an episode represents a single match of a game, in our case this means predicting a complete compression policy $P_e$ for a model $\mathcal{M}$ by using an agent.
Besides predicting a policy $P_e$, an episode comprises the validation of the found policy and the optimization of the used agent, illustrated in Figure~\ref{fig:concept:alg-schema:episode}.
The validation result $V_e$ of this compressed model $M_{P_e}$ consists of accuracy, MACs, BOPs and measured latency and is subsequently used to optimize the agent, which completes the episode.
\begin{figure}[h]
      \centering
      \includegraphics[width=0.49\textwidth]{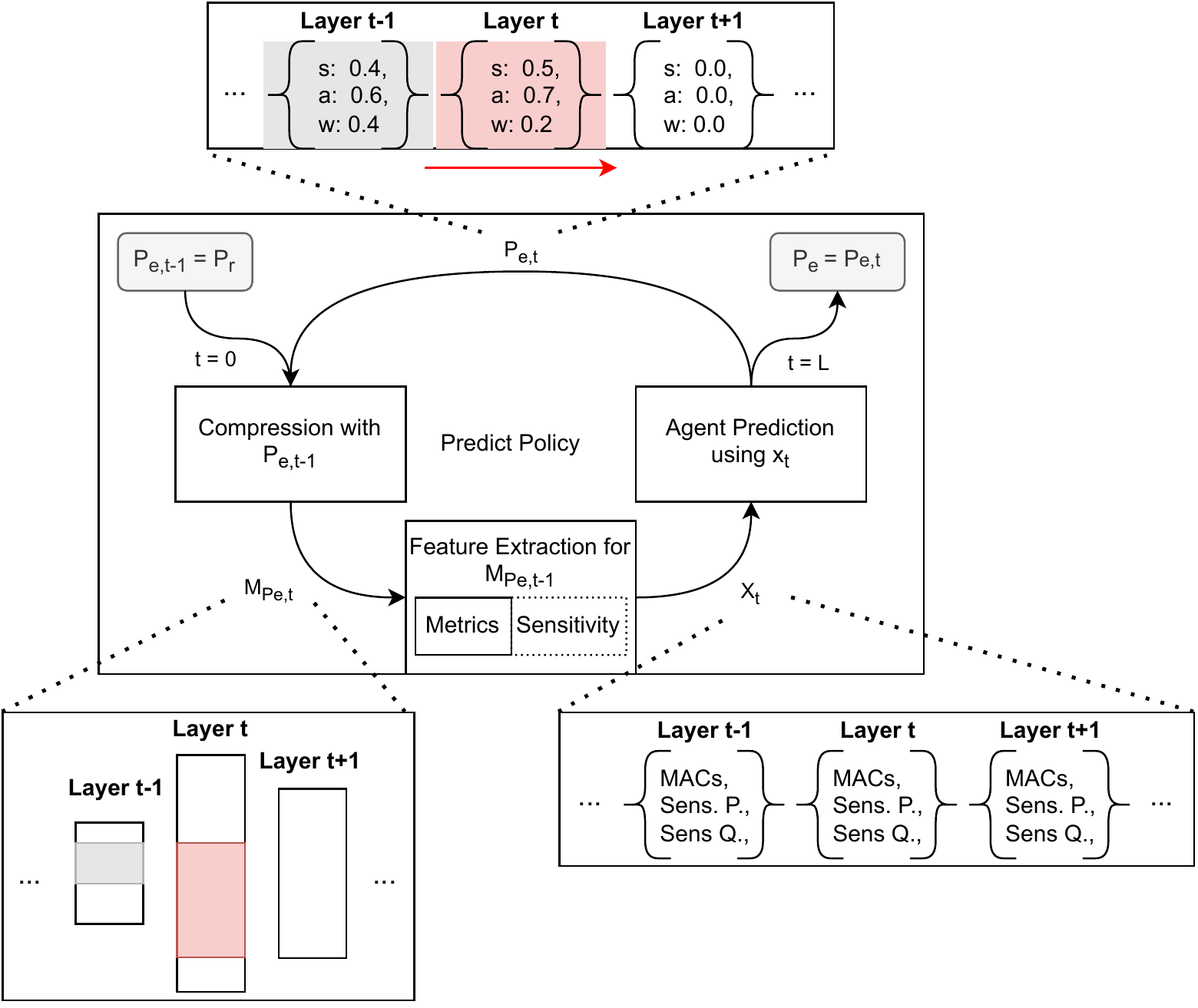}
	   \caption{Overview of the policy prediction cycle. With each step the agent predicts parameters for a single layer.}
	   \label{fig:concept:alg-schema:step} 
\end{figure}

Figure~\ref{fig:concept:alg-schema:step} explains the iterative policy prediction cycle.
With every time step $t$ the algorithm determines the CMPs for all applied compression methods of a single layer.
Starting with a reference policy $P_r$, which is the initial no-compression policy, the algorithm iterates through the model layer by layer, creating a partial policy $P_{e,t}$, which is used to compress the model obtaining $M_{P_{e,t}}$.
The compressed model is required to extract the model features $X_t$ which include metrics like channel count or MACs, and sensitivity results per layer.
Sensitivity results represent a metric measuring the effect of applying a compression method to a single layer.
The agent, using the custom state $\mathbf{s}_t$ created using the model features $X_t$, predicts continuous actions $\mathbf{a}_t \in \mathbb{R}^N$, which then are mapped to the continuous compression parameters $\mathbf{r}$.
Finally, the parameters are mapped to the discrete, implementation and hardware-specific CMPs.

\subsection{Compression Methods}\label{sec:concept:compression}

\paragraph{Pruning}
We implement structured pruning by removing output channels and corresponding weights for convolution layers.
To achieve the given target channel count, which is predicted by the agent, we use the $\ell_1$ strategy~\cite{liPruningFiltersEfficient2017} to identify the channels with least magnitude weights and remove them.
For linear layers, the pruning is implemented analogously by removing output features.

Pruning a layer changes the shape of the output tensor,
thus subsequent layers and operators have to be adjusted accordingly.
Besides the simple case, that the input count of the directly following layer has to be changed, this could yield complex \textit{dependencies}, especially for network architectures with recurrent or residual connections.
We automatically detect such \textit{dependencies} within our algorithm using a specialized library\footnote{Torch-Pruning: \url{https://github.com/VainF/Torch-Pruning}} and do not accept the prediction of pruning parameters for affected layers.

\paragraph{Quantization}
We provide multiple quantization options for applicable layers.
Generally, our algorithm supports mixed-precision quantization with independent bit widths between 1 and 8 bits for activation and weights (\textbf{MIX}), fixed-point 8-bit integer quantization \textbf{INT8}, and no quantization, i.e. single-precision floating point (\textbf{FP32}).
The support for mixed precision quantization differs among hardware targets and is even dependent on concrete layer configurations.
We implement a check for supported quantization actions for each layer and accept only supported quantization actions, analogous to the detection of pruning dependencies.

For accuracy validation during the search we implement \textit{fake quantization}~\cite{gholamiSurveyQuantizationMethods2021}.
For quantization of weight and activation tensors we use uniform quantization with an asymmetric range.
We use dynamic range calibration by selecting minimum and maximum per channel. Formally, mapping value $r$ to its quantized value
\begin{equation}
    Q\left(r\right) = \max{\left(-n, \min{\left(n, \lfloor{s \cdot r - z}\rfloor\right)}\right)},\\
\end{equation}
with $n = 2^b - 1$, scale $s = \frac{n}{x_{max} - x_{min}}$ and offset $z = \lfloor s \cdot x_{min} \rfloor + 2^{b-1}$.
While $b$ is the target bit width and $x_{min}$ and $x_{max}$ defines the range extracted from the tensor.

\subsection{Continuous Actions and Discretization}\label{sec:concept:discretize}
The compression policy $P$ is defined using continuous compression parameters and the action spaces of all our agents are continuous.
We follow the reasoning presented by AMC and HAQ, that a continuous action space allows more fine-grained control and avoids an explosion of the action space while maintaining the order of actions by the resulting compression ratio.
Besides this, we profit from the abstraction of layer-specific details like the concrete channel count.
Thus, the agent predicts an abstract compression ratio per compression method and the action space does not differ per layer.

To apply compression to a model the policy $P$ has to be mapped to discrete CMPs---channel count for pruning and bit widths for quantization.
Thereby we apply an inverse mapping,
\begin{equation}
    d_\nu\left(r\right) = \left\lfloor \left( 1 - r \right) \cdot \nu \right\rfloor + 1,
\end{equation}
where $d_\nu$ maps the compression ratio $r$ to a discrete value using reference $\nu$.
In the case of pruning, the reference is the original channel count of the layer, and for mixed precision quantization, the reference is configurable but at maximum 8 bits.
Because the hardware implementation of some mixed-precision quantization modes require multiples of 8 or 32, we additionally implement the option to round the pruning channel count to a multiple of a fixed value, and thus allow a combined usage of pruning and quantization.

\subsection{Sensitivity Analysis}\label{sec:concept:sensitivity}
To provide some hints to the agent about the effect of compressing a layer we include the results of a sensitivity analysis within the model features $X_t$.
The used sensitivity metric measures the impact of applying a compression method with a defined compression parameter to a single layer.
Thus, it reports how sensitive the overall result of the model is to prune or quantize layer $l$.
We reuse the idea presented by ZeroQ for quantization~\cite{caiZeroQNovelZero2020} and generalize it for a wide range of compression policies.
Subsequently, we measure the distortion introduced by applying compression policy $P$ as
\begin{equation}
    \Omega(P) = \frac{1}{N} \sum^{N}_{j=1} D_{KL}(\mathcal{M}_P(\theta; x_j) \| \mathcal{M}(\theta; x_j))
\end{equation}
where $D_{KL}$ is the Kullback-Leibler divergence measuring the difference of the probability distributions produced by the compressed model compared to the original model.
Thereby, $x_j$ represents one of $N$ samples of the original training data.
To measure the impact of applying a specific CMP configuration to a single layer, we reuse the reference policy $P_r$ and set corresponding parameters only.
Per CMP and layer, a predefined number of sample policies is created.
The complete sensitivity analysis is done upfront the search for all layers.

\subsection{Direct Metric: Hardware Latency}\label{sec:concept:latency}
We use inference latency measured on a specific hardware device as cost metric to optimize for, because latency as a direct metric includes effects specific to the used hardware architecture which could not be characterized by common abstract metrics like MACs or BOPs.
Moreover, the support for quantization is heavily hardware-dependent, some platforms do not support advanced quantization techniques at all.
But even if the techniques are supported, the gained speedup strongly depends on the used implementation and operators~\cite{Klein2021CacheBoundness,sze2020dlprocessormetric}.
By measuring the latency on target devices our algorithm guarantees that the found compression policy could be used in practice and that the found policy is specifically optimized for the concrete device architecture.

We use Apache TVM~\cite{chenTVMAutomatedEndtoend2018} for measuring latency on embedded devices.
TVM is an open-source deep learning compiler that allows to automatically compile, optimize and deploy models to various heterogeneous hardware targets.
This allows us to cross-compile the model during search and instruct an embedded device to perform a latency measurement using the compilation result.
For inference latency testing, we disabled the usage of pre-tuned parameters within the TVM compile step and do not use auto-tuning~\cite{chenLearningOptimizeTensor2019}.
For the experiments in this work we used an ARM Cortex A-72 processor, in place of a huge class of embedded CPUs.
TVM supports convolution and fully-connected bit-serial operators optimized for ARM CPUs with mixed precision~\cite{umurogluOptimizingBitSerialMatrix2019,cowanAutomatingGenerationLow2018,cowanAutomaticGenerationHighperformance2020}.
However, the operator implementation yields some constraints to the configuration of the compressed layer:
For convolution layers the number of input channels must be a full multiple of 32, for output channels a multiple of 8, the spatial output dimension must be at least 2 and depth-wise convolutions are not supported.
For linear layers, the output feature count must be a full multiple of 8.
The mixed-precision compression is restricted to compatible layers. 
Therefore, for joint agents the channel count for pruning has to be rounded to a multiple of 32.

\subsection{Reward Function}\label{sec:concept:reward}

We make use of the \textit{absolute reward function} proposed by \citet{benderCanWeightSharing2020} for the related problem of neural architecture search using reinforcement learning.
The reward function adjusted for our algorithm is therefore,
\begin{equation}
    r\left(P\right) = \textit{acc}_{{M}_P} + \beta \left|\frac{T_{\mathcal{M}_P}}{c \cdot T_{\mathcal{M}}} - 1\right|,
\end{equation}
where $acc_{\mathcal{M}_P}$ is the accuracy of the compressed model, $T_{\mathcal{M}}$  and $T_{\mathcal{M}_P}$ the measured latency of the original and compressed model, respectively.
The hyperparameter $\beta < 0$ is the cost exponent and controls how strong the reward should be reduced when not meeting the target compression rate $c$.
We calculate the reward per episode once for the found policy $P_e$ and assign each time step within the episode the same reward.

We also tried different reward functions, such as \textit{hard exponential reward}~\cite{tanMnasNetPlatformawareNeural2019}, but had similar problems as discussed by \citet{benderCanWeightSharing2020}.

\subsection{Proposed Agents}\label{sec:agents}
We propose three agents to predict compression policies for: quantization, pruning and a joint compression.
While all agents share a common concept and are based on the same DDPG algorithm, the state space $\mathbf{s}_t$ and action space $\mathbf{a}_t$,  the feature-extraction and the mapping of actions to a policy $P$ is action-specific.
Once per episode the compression policy is validated and the calculated reward is shared over all applied transitions.
To reduce the variance, the rewards within the sampled transition batch for optimization are normalized using a moving average.
The states of all agents are normalized by standardization and centralization using mean and variance of the features before feeding them into the agent networks.
As both are unknown we use running estimations updated using seen states, comparable to a batch norm layer.
When starting a new search the agents choose the actions randomly instead of using the actor network for a configurable number of episodes.
These \emph{warm-up episodes} are required to fill the replay buffers with enough transitions before executing the first optimization of the agent.
To add exploration noise we sample each action from a truncated normal,
\begin{equation}
    a_t' \sim \mathcal{N}_{trunc}\left(\mu(s_t|\theta^{\mu}),\sigma^2, 0, 1\right),
\end{equation}
where $\mu(s_t|\theta^{\mu})$ is the original prediction of the actor network of the corresponding agent.
The used noise derivation $\sigma$ decays exponentially, therefore the exploration noise decreases each episode.
With that, the first episodes of a search assemble the exploration phase which smoothly blends into the exploitation phase of the algorithm.
We use an initial noise derivation of $\sigma = 0.5$ and a decay rate of $0.95$.

The actor and critic networks used for all agents consist of two hidden linear layers with 400 and 300 features.
All actions predicted by the agents are limited to $[0,1]$ by applying a Sigmoid activation function to the output layers of the actor networks.
We set the discount factor $\gamma$ within the Bellman equation for Q-learning to $0.99$, the factor controls the horizon of the expected reward calculation.
For optimization of the actor and critic networks we use the Adam optimizer~\cite{kingmaAdamMethodStochastic2015} with a learning rate of $0.0001$ for the actor network and $0.001$ for the critic network. 
For both we use $\beta_1=0.9$ and $\beta_2=0.999$.
The batch size for the agent optimization is 128.
We use a replay-buffer-size of 2000, but since the number of transitions per episode is agent and model dependent, the real number of episodes in the buffer differs.

\subsection{Quantization Implementation Details}
\paragraph{Selection of Quantization Method}\label{sec:agents:mapping}
We support three different quantization methods, which can be applied layer-wise.
We select the quantization method by applying thresholds based on the predicted actions.
If activation $a_a$ or weight $a_w$ action exceeds threshold $t_{mix} = 0.5$ \textbf{MIX}  quantization, otherwise, if one of them exceeds $t_{int8} = 0.2$ \textbf{INT8} quantization, otherwise \textbf{FP32} is used.
For layers which do not support mixed precision quantization the agent selects the \textbf{INT8} option instead.
The mixed precision quantization requires continuous compression parameters, thus we scale the actions $a_a$ and $a_w$ to the compression parameters $r_a, r_w \in [0,1]$ with:
\begin{equation}
    r_i = \max\left(\min\left(\frac{a_i-t_{mix}}{1-t_{mix}}, 0\right), 1\right).
\end{equation}

\paragraph{Exploration Range}
The implementation supports limiting the maximum bit widths for the \textbf{MIX} quantization option.
For \textit{ResNet18} we validated that bit widths with more than 6 bits lead to slower inference times for the used bit-serial operation compared to the \textbf{INT8} option.

Additionally, we discovered that the TVM compile time spikes drastically when bit-serial operations with high bit widths are used.
Therefore, we limit the maximum bit width for the \textbf{MIX} option to 6 bits for all our experiments to avoid unnecessary long exploration phases and shorten the search time significantly.

\section{Experiments}
We evaluated the proposed algorithm using the three agents with a ResNet18~\cite{heDeepResidualLearning2016} trained on the CIFAR-10 dataset~\cite{krizhevskyLearningMultipleLayers2009}.
We split a custom validation set from the train data set and use it for accuracy validation and sensitivity analysis.
For all experiments, we used a Raspberry Pi Model 4B with an ARM Cortex A-72 processor as hardware target to measure inference latency.
For the quantization agent, we ran 310 episodes per experiment and 410 episodes for the pruning and the joint agent.
We included 10 warm-up episodes at the beginning of each search and used for all experiments in the reward function a cost exponent of $\beta$\,=\,-3.0.
Reported accuracies are test accuracies of the compressed and for 30 epochs retrained models.

\subsection{Comparing Agent Policies}
The goal of the experiment is to validate the basic functionality of the algorithm and the three agents.
Therefore, we evaluated policy searches using all three agents with various target compression rates.

\paragraph{General Performance}
Table~\ref{tab:experiments:agents} shows, that comparing the performance of the three different agents with a compression ratio of c\,=\,0.3, all agents are successful at compressing the model and reducing the latency to the aimed 30\,\% of the original model.
This illustrates that every single agent is suitable to find optimized compression policies using the available methods with optimized, layer-specific compression ratios.
While for less challenging target compression rates, all agents can find optimized compression policies reaching target latency without notable loss in accuracy, in extreme conditions (e.g. Table~\ref{tab:experiments:agents}, c\,=\,0.2) the quantization agent is forced to use extreme small bit widths to reach the target compression ratio, which leads to a huge loss in accuracy.
We suspect that for such extreme 1-bit quantization, if possible at all, advanced methods are required to sustain accuracy.

While the pruning agent reduces the amount of MACs most and the quantization agent is the most effective in minimizing BOPs, the joint agent balances both compression methods.
Since the desired latency reduction is a preset parameter, the agents try to use all resources in this indirect budget to preserve accuracy.
The joint agent can exploit quantization and pruning combined and can achieve the latency reduction with less aggressive usage of both methods with best conservation of accuracy.

\tabcolsep=0.05cm
\begin{table}[!t]\renewcommand{\arraystretch}{1.2} 
	\caption{Compressed model performance per agent with target compression ratio $c$ }
	\label{tab_agents_overview}
	\centering
	\small	
	\begin{tabular}{l|c|l|r|r|r}
	\bfseries Method & \bfseries c & \multicolumn{1}{c|}{\bfseries MACs} & \multicolumn{1}{c|}{\bfseries BOPs} &  \multicolumn{1}{c|}{\bfseries Latency} & \multicolumn{1}{c}{\bfseries Accuracy} \\ \hline
	Uncompressed 			& 							& $4.75 \cdot 10^{10}$ 	&  $4.86 \cdot 10^{13}$ 		&  $330$\,ms 		&  	$93.0$\,\% \\ 
	\hline
	Pruning Agent			& \multirow{3}{*}{0.3} & $1.42 \cdot 10^{10}$ 	&  $1.45 \cdot 10^{13}$ 		&  $98$\,ms 		& 	$93.0$\,\% \\
	Quantization A.		&							& $4.75 \cdot 10^{10}$ 	&  $8.23 \cdot 10^{11}$ 		&  $98$\,ms 		&  	$92.5$\,\% \\
	Joint Agent			&							& $4.35 \cdot 10^{10}$ 	&  $9.42 \cdot 10^{11}$ 		&  $99$\,ms 		& 	$93.2$\,\% \\
	\hline
	Pruning Agent			& \multirow{3}{*}{0.2}	& $9.24 \cdot 10^{9}$ 	&  $9.45 \cdot 10^{12}$ 		&  $66$\,ms 		& 	$92.4$\,\% \\
	Quantization A.		&							& $4.75 \cdot 10^{10}$ 	&  $4.01 \cdot 10^{11}$ 		&  $57$\,ms 		&  	$45.0$\,\% \\
	Joint Agent			&							& $2.82 \cdot 10^{10}$ 	&  $6.74 \cdot 10^{11}$ 		&  $64$\,ms 		& 	$92.8$\,\% \\
	\end{tabular}
	\label{tab:experiments:agents}
\end{table}

\begin{figure}
    \begin{subfigure}{0.49\textwidth}
        \centering
        \includegraphics[width=\textwidth]{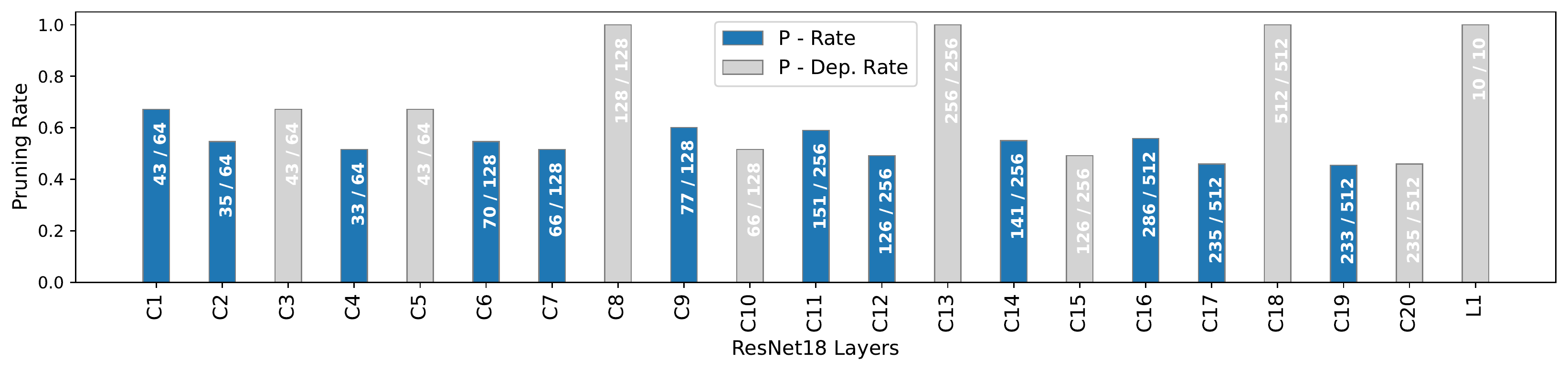}
        \caption{Pruning Agent}
        \label{fig:exp01a:prune-discrete}
    \end{subfigure}
    \begin{subfigure}{0.49\textwidth}
        \centering
        \includegraphics[width=\textwidth]{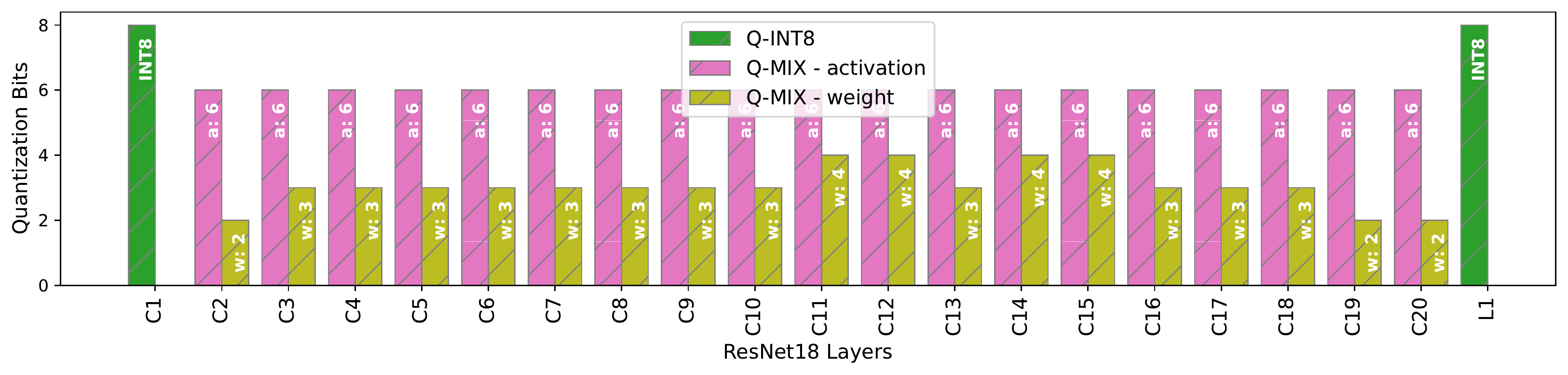}
        \caption{Quantization Agent}
        \label{fig:exp01b:quantize-discrete}
    \end{subfigure}
    \begin{subfigure}{0.49\textwidth}
        \centering
        \includegraphics[width=\textwidth]{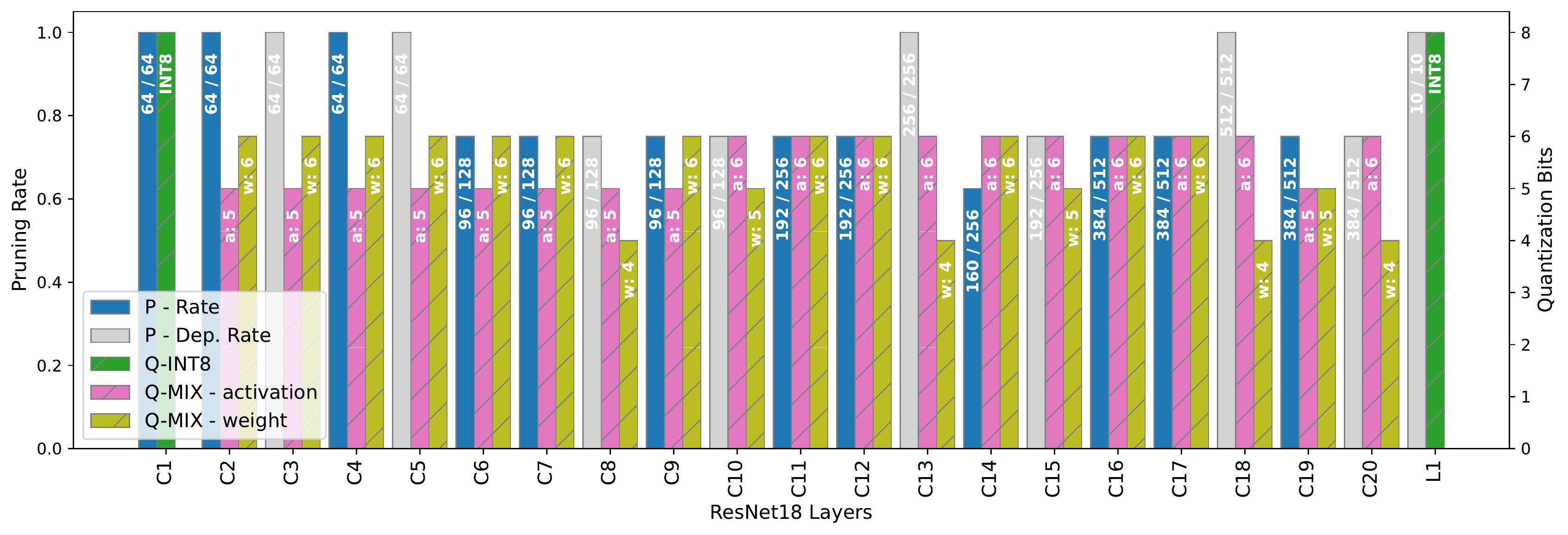}
        \caption{Joint Agent}
        \label{fig:exp01c:pq-discrete}
    \end{subfigure}
    \setlength{\belowcaptionskip}{-10pt}
    \caption{Predicted compression policies $P$ of the pruning, quantization and joint agent. With a target compression rate of c\,=\,0.3. Bar labels indicate remaining channels for pruning, and bit width for activations and weights, respectively.}\label{fig:exp01}
\end{figure}

\paragraph{Policy Analysis}
To compare the policies of the different agents, we used a less challenging compression rate of $c=0.3$, such that the observed policies produce comparable accuracies.
The pruning agent seems to prune all layers---except for the first---almost equally, illustrated in Figure~\ref{fig:exp01a:prune-discrete}, with a minor tendency to prune latter layers more.
The other exceptional type of layers, the gray-colored layers, depend on other layers and could not be pruned independently.

The quantization agent in contrast, as illustrated in Figure~\ref{fig:exp01b:quantize-discrete}, varies the quantization bit widths more across all layers.
The usage of \textbf{INT8} quantization for the first and last layer is induced by the constraints for using the \textbf{MIX} quantization and hence no explicit decision of the agent.
A slight trend towards smaller bit widths for first and last layers is detectable, at the same time layers in the middle of the network have the largest bit widths.
The agent quantizes weights much stronger than activations, which is also a common pattern in hand-tuned quantized models, since often the activations are more sensitive to quantization noise~\cite{zhou2016dorefa,zhu2017ttq,schindler2018}.

The joint agent follows a mixed pattern.
Figure~\ref{fig:exp01c:pq-discrete} shows that it quantizes activations up to 5 bits for activations and weights up to 4 bits, overall less aggressive than the quantization agent.
\textbf{INT8} quantization is again only used for layers without stronger alternatives.
In contrast to the pruning agent this joint agent does not prune the first layers and uses pruning, more limited, probably also due to the restriction of pruning only multiples of 32 which are rather large parts of the channel-wise small first layers.
Due to the computational savings quantization, less pruning is required and vice versa.
Overall the joint agent has a larger action space and use this freedom for a more balanced compression.

\subsection{Variation of Target Compression Rate $c$}
We do not enforce the target compression rate $c$ by overriding or clipping actions like related approaches~\cite{heAMCAutoMLModel2019,wangHAQHardwareAwareAutomated2019}.
Instead, we include the target within the reward function.
Within the following experiment, we vary the target compression rate and test thereby if the agents are capable to predict policies matching the given resource budget.
\begin{figure}
    \centering
    \includegraphics[width=0.49\textwidth]{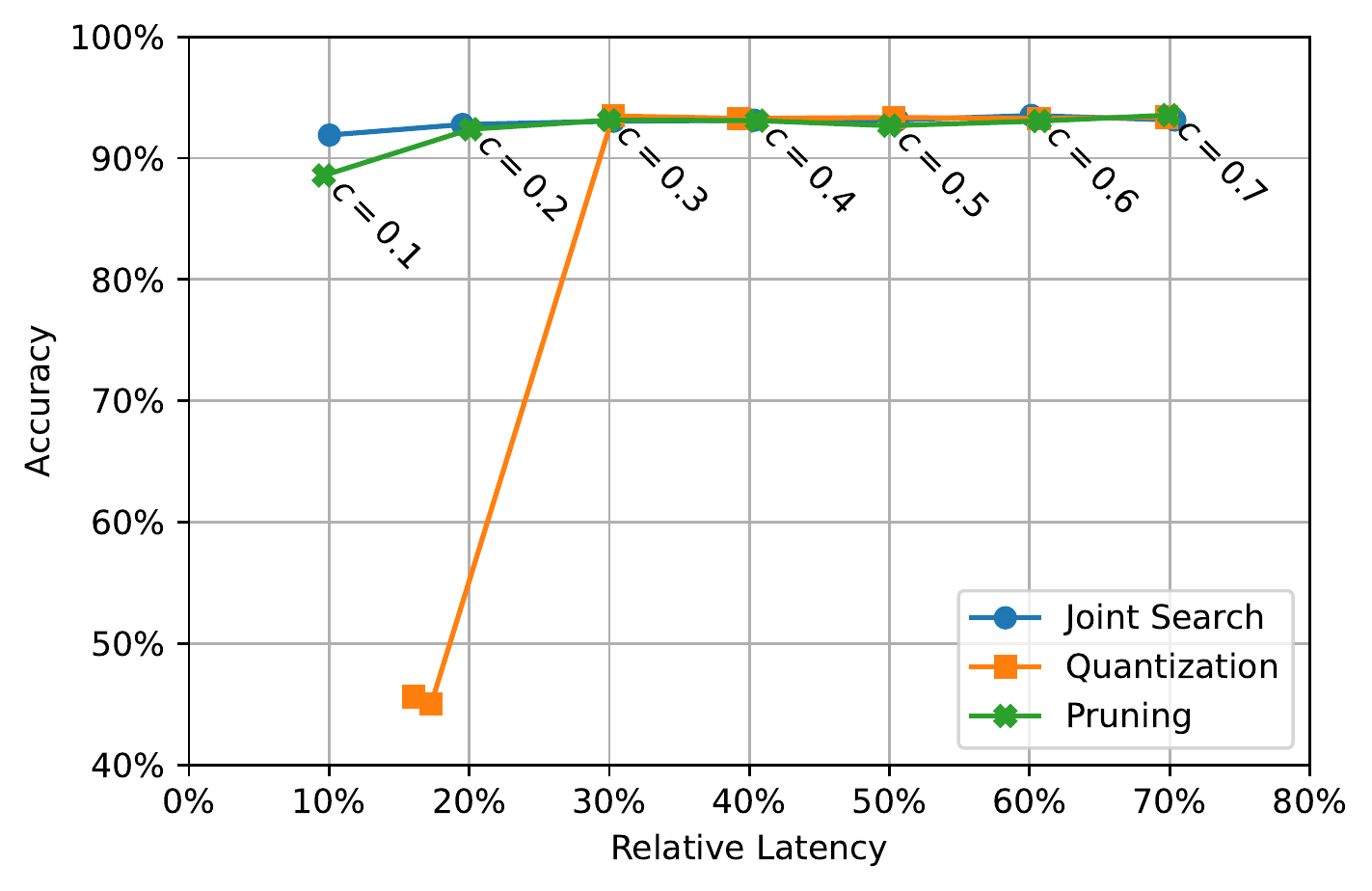}
    \label{fig:exp02:frontier-full}
    \setlength{\abovecaptionskip}{-5pt}
    \setlength{\belowcaptionskip}{-10pt}
    \caption{Comparing the accuracy and relative latency of the agents with various target compression rates $c$.}\label{fig:exp02:frontier}
\end{figure} 

Figure~\ref{fig:exp02:frontier} shows for each tested compression ratio $c \in \{0.1, 0.2, 0.3, 0.4, 0.5, 0.6, 0.7\}$ the achieved relative latency and accuracy.
For the most challenging compression ratios the quantization agent is forced to use extreme small bit widths, finally failing to achieve the target latency, accompanied by a huge loss in accuracy.
This demonstrates that if we exceed the limit to which a model can be compressed with a specific compression method, the agent finally fails to find a useful policy, overwhelmed by balancing too demanding latency and accuracy constraints.
For the other two agents, a decrease in accuracy with increasing compression rate is observable too, however, the results are still in an acceptable range.

Despite the extreme quantization case, the policies found are observed within 5 percent points of the target latency. 
This demonstrates that the control of the resource budget by the reward function is quite effective.
The hyperparameter $\beta$ even provides the possibility to relax or strengthen the constraint.
In addition, we consider policies with latencies smaller than the given target as acceptable, although the used reward also penalizes these.
The results show that an automatic compression search using measured inference latency is suitable.

The joint search resulted in the best accuracy for small compression rate targets.
Although the difference in accuracy compared to the pruning agent is quite small, for these compression targets a superiority is detectable.
Combined with the sharp decrease for the quantization agent this illustrates the value of a combined search using both compression methods.
A detailed analysis of the policies predicted by the agents underlines that the joint agent constantly applies less restrictive compression using both methods in a balanced manner.

Other ablation studies include a demonstration that a concurrent joint policy search is balancing better than a sequential series of pruning and quantization searches, or variations of such sequential approaches. Furthermore, another study shows that the sensitivity information enables the agents to exploit heterogeneity in compression for the different layers better, thereby compressing the most resilient layers most. In particular considering scalable model architectures, one can assume an increasing benefit of the sensitivity information. Short summaries of these ablation studies can be found in the appendix.

\section{Summary and Outlook} 
We introduced an algorithmic concept called "Galen" for the automatic compression of neural networks using reinforcement learning, consisting of an automated framework and three proposed agents for quantization, pruning, and joint compression, respectively.
Contrary to other approaches, Galen validates the compressed model by deploying and benchmarking on a real-world embedded system, using code generation with support for sparse and quantized operators.
With that, we use real inference latency as our optimization target within the search algorithm, and predict compression policies specific to the selected target and existing hardware constraints.
While the algorithm itself is generic and extendable to further compression methods, we support pruning and quantization, and notably joint pruning and quantization---with support for different quantization types.
Thereby, we demonstrate that it is sufficient to specify the inference-latency budget as constraint within the reward function.

For the first results of ResNet18 on CIFAR-10 we can report nearly perfect compression:
Using our joint agent we compressed the model to 20\,\% of the original latency while achieving an accuracy of 92.8\,\%.
For compression to 30\,\% of the original latency, we fully conserved the original accuracy.  
With that, we also infer the obvious next steps, that validation of the algorithm and the proposed agents on more complex data sets and various model architectures is required.

By comparing the joint agent to the pruning and quantization agent, we can replicate the known result that the combination of both compression methods is very effective to achieve high compression rates with top accuracies~\cite{hanDeepCompressionCompressing2016}.
The detailed analysis of predicted compression policies of the different agents leads to the insight, that a joint agent---guided by a sensitivity metric---can balance the impact of compression over different layers and compression methods.

Overall these are the first results and proof of concept, with great opportunities for further extensions.
Very promising and unique would be the integration of detailed, layer-wise hardware feedback.
Performance counters---providing for example the cache miss rate---could be evaluated to guide the agents, not only by sensitivity, but also by hardware performance metrics.

\section*{Acknowledgment}
This work is part of the COMET program within the K2 Center “Integrated Computational Material, Process and Product Engineering (IC-MPPE)” (Project No 886385), and supported by the Austrian Federal Ministries for Climate Action, Environment, Energy, Mobility, Innovation and Technology (BMK) and for Labour and Economy (BMAW), represented by the Austrian Research Promotion Agency (FFG), and the federal states of Styria, Upper Austria and Tyrol.
\bibliography{references}

\begin{thebibliography}{37}
\providecommand{\natexlab}[1]{#1}

\bibitem[{Baskin et~al.(2021)Baskin, Liss, Schwartz, Zheltonozhskii, Giryes,
  Bronstein, and Mendelson}]{Baskin2021bops}
Baskin, C.; Liss, N.; Schwartz, E.; Zheltonozhskii, E.; Giryes, R.; Bronstein,
  A.~M.; and Mendelson, A. 2021.
\newblock UNIQ: Uniform Noise Injection for Non-Uniform Quantization of Neural
  Networks.
\newblock \emph{ACM Trans. Comput. Syst.}, 37(1–4).

\bibitem[{Bender et~al.(2020)Bender, Liu, Chen, Chu, Cheng, Kindermans, and
  Le}]{benderCanWeightSharing2020}
Bender, G.; Liu, H.; Chen, B.; Chu, G.; Cheng, S.; Kindermans, P.-J.; and Le,
  Q.~V. 2020.
\newblock Can {{Weight Sharing Outperform Random Architecture Search}}? {{An
  Investigation With TuNAS}}.
\newblock In \emph{Proceedings of the {{IEEE}}/{{CVF Conference}} on {{Computer
  Vision}} and {{Pattern Recognition}}}, 14323--14332.

\bibitem[{Cai et~al.(2020)Cai, Yao, Dong, Gholami, Mahoney, and
  Keutzer}]{caiZeroQNovelZero2020}
Cai, Y.; Yao, Z.; Dong, Z.; Gholami, A.; Mahoney, M.~W.; and Keutzer, K. 2020.
\newblock {{ZeroQ}}: {{A Novel Zero Shot Quantization Framework}}.
\newblock In \emph{2020 {{IEEE}}/{{CVF Conference}} on {{Computer Vision}} and
  {{Pattern Recognition}} ({{CVPR}})}, 13166--13175. {Seattle, WA, USA}:
  {IEEE}.
\newblock ISBN 978-1-72817-168-5.

\bibitem[{Chen et~al.(2018)Chen, Moreau, Jiang, Zheng, Yan, Cowan, Shen, Wang,
  Hu, Ceze, Guestrin, and Krishnamurthy}]{chenTVMAutomatedEndtoend2018}
Chen, T.; Moreau, T.; Jiang, Z.; Zheng, L.; Yan, E.; Cowan, M.; Shen, H.; Wang,
  L.; Hu, Y.; Ceze, L.; Guestrin, C.; and Krishnamurthy, A. 2018.
\newblock {{TVM}}: {{An}} Automated End-to-End Optimizing Compiler for Deep
  Learning.
\newblock In \emph{13th {{USENIX}} Conference on Operating Systems Design and
  Implementation}, {{OSDI}}. {Carlsbad, CA, USA}.
\newblock ISBN 978-1-931971-47-8.

\bibitem[{Chen et~al.(2019)Chen, Zheng, Yan, Jiang, Moreau, Ceze, Guestrin, and
  Krishnamurthy}]{chenLearningOptimizeTensor2019}
Chen, T.; Zheng, L.; Yan, E.; Jiang, Z.; Moreau, T.; Ceze, L.; Guestrin, C.;
  and Krishnamurthy, A. 2019.
\newblock Learning to {{Optimize Tensor Programs}}.
\newblock arXiv:1805.08166.

\bibitem[{Cowan et~al.(2020)Cowan, Moreau, Chen, Bornholt, and
  Ceze}]{cowanAutomaticGenerationHighperformance2020}
Cowan, M.; Moreau, T.; Chen, T.; Bornholt, J.; and Ceze, L. 2020.
\newblock Automatic Generation of High-Performance Quantized Machine Learning
  Kernels.
\newblock In \emph{Proceedings of the 18th {{ACM}}/{{IEEE International
  Symposium}} on {{Code Generation}} and {{Optimization}}}, 305--316. {San
  Diego CA USA}: {ACM}.
\newblock ISBN 978-1-4503-7047-9.

\bibitem[{Cowan et~al.(2018)Cowan, Moreau, Chen, and
  Ceze}]{cowanAutomatingGenerationLow2018}
Cowan, M.; Moreau, T.; Chen, T.; and Ceze, L. 2018.
\newblock Automating Generation of Low Precision Deep Learning Operators.
\newblock \emph{CoRR}, abs/1810.11066.

\bibitem[{Dong et~al.(2020)Dong, Yao, Arfeen, Gholami, Mahoney, and
  Keutzer}]{dongHAWQV2HessianAware2020}
Dong, Z.; Yao, Z.; Arfeen, D.; Gholami, A.; Mahoney, M.~W.; and Keutzer, K.
  2020.
\newblock {{HAWQ-V2}}: {{Hessian}} Aware Trace-{{Weighted}} Quantization of
  Neural Networks.
\newblock In Larochelle, H.; Ranzato, M.; Hadsell, R.; Balcan, M.; and Lin, H.,
  eds., \emph{Advances in Neural Information Processing Systems}, volume~33,
  18518--18529. {Curran Associates, Inc.}

\bibitem[{Dong et~al.(2019)Dong, Yao, Gholami, Mahoney, and
  Keutzer}]{dongHAWQHessianAWare2019}
Dong, Z.; Yao, Z.; Gholami, A.; Mahoney, M.; and Keutzer, K. 2019.
\newblock {{HAWQ}}: {{Hessian AWare Quantization}} of {{Neural Networks With
  Mixed-Precision}}.
\newblock In \emph{2019 {{IEEE}}/{{CVF International Conference}} on {{Computer
  Vision}} ({{ICCV}})}, 293--302. {Seoul, Korea (South)}: {IEEE}.
\newblock ISBN 978-1-72814-803-8.

\bibitem[{Elthakeb et~al.(2020)Elthakeb, Pilligundla, Mireshghallah,
  Yazdanbakhsh, and Esmaeilzadeh}]{elthakebReLeQReinforcementLearning2020}
Elthakeb, A.~T.; Pilligundla, P.; Mireshghallah, F.; Yazdanbakhsh, A.; and
  Esmaeilzadeh, H. 2020.
\newblock {{ReLeQ}} : {{A Reinforcement Learning Approach}} for {{Automatic
  Deep Quantization}} of {{Neural Networks}}.
\newblock \emph{IEEE Micro}, 40(5): 37--45.

\bibitem[{Gholami et~al.(2021)Gholami, Kim, Dong, Yao, Mahoney, and
  Keutzer}]{gholamiSurveyQuantizationMethods2021}
Gholami, A.; Kim, S.; Dong, Z.; Yao, Z.; Mahoney, M.~W.; and Keutzer, K. 2021.
\newblock A {{Survey}} of {{Quantization Methods}} for {{Efficient Neural
  Network Inference}}.
\newblock \emph{arXiv:2103.13630 [cs]}.

\bibitem[{Han, Mao, and Dally(2016)}]{hanDeepCompressionCompressing2016}
Han, S.; Mao, H.; and Dally, W.~J. 2016.
\newblock Deep {{Compression}}: {{Compressing Deep Neural Network}} with
  {{Pruning}}, {{Trained Quantization}} and {{Huffman Coding}}.
\newblock In Bengio, Y.; and LeCun, Y., eds., \emph{4th {{International
  Conference}} on {{Learning Representations}}, {{ICLR}} 2016, {{San Juan}},
  {{Puerto Rico}}, {{May}} 2-4, 2016, {{Conference Track Proceedings}}}.

\bibitem[{He et~al.(2016)He, Zhang, Ren, and Sun}]{heDeepResidualLearning2016}
He, K.; Zhang, X.; Ren, S.; and Sun, J. 2016.
\newblock Deep {{Residual Learning}} for {{Image Recognition}}.
\newblock In \emph{2016 {{IEEE Conference}} on {{Computer Vision}} and
  {{Pattern Recognition}} ({{CVPR}})}, 770--778.

\bibitem[{He et~al.(2019)He, Lin, Liu, Wang, Li, and
  Han}]{heAMCAutoMLModel2019}
He, Y.; Lin, J.; Liu, Z.; Wang, H.; Li, L.-J.; and Han, S. 2019.
\newblock {{AMC}}: {{AutoML}} for {{Model Compression}} and {{Acceleration}} on
  {{Mobile Devices}}.
\newblock \emph{arXiv:1802.03494 [cs]}.

\bibitem[{He, Zhang, and Sun(2017)}]{heChannelPruningAccelerating2017}
He, Y.; Zhang, X.; and Sun, J. 2017.
\newblock Channel {{Pruning}} for {{Accelerating Very Deep Neural Networks}}.
\newblock In \emph{2017 {{IEEE International Conference}} on {{Computer
  Vision}} ({{ICCV}})}, 1398--1406. {Venice}: {IEEE}.
\newblock ISBN 978-1-5386-1032-9.

\bibitem[{Jacob et~al.(2018)Jacob, Kligys, Chen, Zhu, Tang, Howard, Adam, and
  Kalenichenko}]{jacobQuantizationTrainingNeural2018}
Jacob, B.; Kligys, S.; Chen, B.; Zhu, M.; Tang, M.; Howard, A.; Adam, H.; and
  Kalenichenko, D. 2018.
\newblock Quantization and {{Training}} of {{Neural Networks}} for {{Efficient
  Integer-Arithmetic-Only Inference}}.
\newblock In \emph{Proceedings of the {{IEEE Conference}} on {{Computer
  Vision}} and {{Pattern Recognition}}}, 2704--2713.

\bibitem[{Kingma and Ba(2015)}]{kingmaAdamMethodStochastic2015}
Kingma, D.~P.; and Ba, J. 2015.
\newblock Adam: {{A}} Method for Stochastic Optimization.
\newblock In Bengio, Y.; and LeCun, Y., eds., \emph{3rd International
  Conference on Learning Representations, {{ICLR}} 2015, San Diego, {{CA}},
  {{USA}}, May 7-9, 2015, Conference Track Proceedings}.

\bibitem[{Klein et~al.(2021)Klein, Gratl, M{\"u}cke, and
  Fr{\"o}ning}]{Klein2021CacheBoundness}
Klein, B.; Gratl, C.; M{\"u}cke, M.; and Fr{\"o}ning, H. 2021.
\newblock Understanding Cache Boundness of {{ML}} Operators on {{ARM}}
  Processors.
\newblock In \emph{3rd Workshop on Accelerated Machine Learning ({{AccML}})}.
  {HiPEAC 2021 Conference}.

\bibitem[{Krizhevsky, Hinton
  et~al.(2009)}]{krizhevskyLearningMultipleLayers2009}
Krizhevsky, A.; Hinton, G.; et~al. 2009.
\newblock Learning Multiple Layers of Features from Tiny Images.

\bibitem[{Li et~al.(2017)Li, Kadav, Durdanovic, Samet, and
  Graf}]{liPruningFiltersEfficient2017}
Li, H.; Kadav, A.; Durdanovic, I.; Samet, H.; and Graf, H.~P. 2017.
\newblock Pruning {{Filters}} for {{Efficient ConvNets}}.
\newblock \emph{arXiv:1608.08710 [cs]}.

\bibitem[{Lillicrap et~al.(2019)Lillicrap, Hunt, Pritzel, Heess, Erez, Tassa,
  Silver, and Wierstra}]{lillicrapContinuousControlDeep2019}
Lillicrap, T.~P.; Hunt, J.~J.; Pritzel, A.; Heess, N.; Erez, T.; Tassa, Y.;
  Silver, D.; and Wierstra, D. 2019.
\newblock Continuous Control with Deep Reinforcement Learning.
\newblock \emph{arXiv:1509.02971 [cs, stat]}.

\bibitem[{Lin et~al.(2020)Lin, Ji, Zhang, Zhang, Wu, and
  Tian}]{linChannelPruningAutomatic2020}
Lin, M.; Ji, R.; Zhang, Y.; Zhang, B.; Wu, Y.; and Tian, Y. 2020.
\newblock Channel {{Pruning}} via {{Automatic Structure Search}}.
\newblock In \emph{Proceedings of the {{Twenty-Ninth International Joint
  Conference}} on {{Artificial Intelligence}}}, 673--679. {Yokohama, Japan}:
  {International Joint Conferences on Artificial Intelligence Organization}.
\newblock ISBN 978-0-9992411-6-5.

\bibitem[{Liu et~al.(2020)Liu, Ma, Xu, Wang, Tang, and
  Ye}]{liuAutoCompressAutomaticDNN2020}
Liu, N.; Ma, X.; Xu, Z.; Wang, Y.; Tang, J.; and Ye, J. 2020.
\newblock {{AutoCompress}}: {{An}} Automatic {{DNN}} Structured Pruning
  Framework for Ultra-High Compression Rates.
\newblock \emph{Proceedings of the AAAI Conference on Artificial Intelligence},
  34(04): 4876--4883.

\bibitem[{Lou et~al.(2020)Lou, Guo, Kim, Liu, and
  Jiang}]{louAutoQAutomatedKernelwise2020}
Lou, Q.; Guo, F.; Kim, M.; Liu, L.; and Jiang, L. 2020.
\newblock {{AutoQ}}: {{Automated}} Kernel-Wise Neural Network Quantization.
\newblock In \emph{8th International Conference on Learning Representations,
  {{ICLR}} 2020, Addis Ababa, Ethiopia, April 26-30, 2020}. {OpenReview.net}.

\bibitem[{Schindler et~al.(2018)Schindler, Z{\"o}hrer, Pernkopf, and
  Fr{\"o}ning}]{schindler2018}
Schindler, G.; Z{\"o}hrer, M.; Pernkopf, F.; and Fr{\"o}ning, H. 2018.
\newblock Towards efficient forward propagation on resource-constrained
  systems.
\newblock In \emph{Joint European Conference on Machine Learning and Knowledge
  Discovery in Databases}, 426--442. Springer.

\bibitem[{Sze et~al.(2020)Sze, Chen, Yang, and Emer}]{sze2020dlprocessormetric}
Sze, V.; Chen, Y.-H.; Yang, T.-J.; and Emer, J.~S. 2020.
\newblock How to Evaluate Deep Neural Network Processors: {{TOPS}}/{{W}}
  (Alone) Considered Harmful.
\newblock \emph{IEEE Solid-State Circuits Magazine}, 12(3): 28--41.

\bibitem[{Tan et~al.(2019)Tan, Chen, Pang, Vasudevan, Sandler, Howard, and
  Le}]{tanMnasNetPlatformawareNeural2019}
Tan, M.; Chen, B.; Pang, R.; Vasudevan, V.; Sandler, M.; Howard, A.; and Le,
  Q.~V. 2019.
\newblock {{MnasNet}}: {{Platform-aware}} Neural Architecture Search for
  Mobile.
\newblock In \emph{Proceedings of the {{IEEE}}/{{CVF}} Conference on Computer
  Vision and Pattern Recognition ({{CVPR}})}.

\bibitem[{Tung and Mori(2018)}]{tungCLIPQDeepNetwork2018}
Tung, F.; and Mori, G. 2018.
\newblock {{CLIP-Q}}: {{Deep Network Compression Learning}} by {{In-parallel
  Pruning-Quantization}}.
\newblock In \emph{2018 {{IEEE}}/{{CVF Conference}} on {{Computer Vision}} and
  {{Pattern Recognition}}}, 7873--7882.

\bibitem[{Umuroglu et~al.(2019)Umuroglu, Conficconi, Rasnayake, Preusser, and
  Sj{\"a}lander}]{umurogluOptimizingBitSerialMatrix2019}
Umuroglu, Y.; Conficconi, D.; Rasnayake, L.; Preusser, T.~B.; and
  Sj{\"a}lander, M. 2019.
\newblock Optimizing {{Bit-Serial Matrix Multiplication}} for {{Reconfigurable
  Computing}}.
\newblock \emph{ACM Transactions on Reconfigurable Technology and Systems},
  12(3).

\bibitem[{Wang et~al.(2019)Wang, Liu, Lin, Lin, and
  Han}]{wangHAQHardwareAwareAutomated2019}
Wang, K.; Liu, Z.; Lin, Y.; Lin, J.; and Han, S. 2019.
\newblock {{HAQ}}: {{Hardware-Aware Automated Quantization With Mixed
  Precision}}.
\newblock In \emph{2019 {{IEEE}}/{{CVF Conference}} on {{Computer Vision}} and
  {{Pattern Recognition}} ({{CVPR}})}, 8604--8612. {Long Beach, CA, USA}:
  {IEEE}.
\newblock ISBN 978-1-72813-293-8.

\bibitem[{Wang et~al.(2020)Wang, Wang, Cai, Lin, Liu, Wang, Lin, and
  Han}]{wangAPQJointSearch2020}
Wang, T.; Wang, K.; Cai, H.; Lin, J.; Liu, Z.; Wang, H.; Lin, Y.; and Han, S.
  2020.
\newblock {{APQ}}: {{Joint Search}} for {{Network Architecture}}, {{Pruning}}
  and {{Quantization Policy}}.
\newblock In \emph{2020 {{IEEE}}/{{CVF Conference}} on {{Computer Vision}} and
  {{Pattern Recognition}} ({{CVPR}})}, 2075--2084. {Seattle, WA, USA}: {IEEE}.
\newblock ISBN 978-1-72817-168-5.

\bibitem[{Wang, Lu, and Blankevoort(2020)}]{wangDifferentiableJointPruning2020}
Wang, Y.; Lu, Y.; and Blankevoort, T. 2020.
\newblock Differentiable {{Joint Pruning}} and {{Quantization}} for {{Hardware
  Efficiency}}.
\newblock In Vedaldi, A.; Bischof, H.; Brox, T.; and Frahm, J.-M., eds.,
  \emph{Computer {{Vision}} \textendash{} {{ECCV}} 2020}, Lecture {{Notes}} in
  {{Computer Science}}, 259--277. {Cham}: {Springer International Publishing}.
\newblock ISBN 978-3-030-58526-6.

\bibitem[{Yang et~al.(2020)Yang, Gui, Zhu, and
  Liu}]{yangAutomaticNeuralNetwork2020}
Yang, H.; Gui, S.; Zhu, Y.; and Liu, J. 2020.
\newblock Automatic Neural Network Compression by Sparsity-Quantization Joint
  Learning: {{A}} Constrained Optimization-Based Approach.
\newblock In \emph{Proceedings of the {{IEEE}}/{{CVF}} Conference on Computer
  Vision and Pattern Recognition ({{CVPR}})}.

\bibitem[{Yang et~al.(2018)Yang, Howard, Chen, Zhang, Go, Sandler, Sze, and
  Adam}]{yangNetAdaptPlatformawareNeural2018}
Yang, T.-J.; Howard, A.; Chen, B.; Zhang, X.; Go, A.; Sandler, M.; Sze, V.; and
  Adam, H. 2018.
\newblock {{NetAdapt}}: {{Platform-aware}} Neural Network Adaptation for Mobile
  Applications.
\newblock In \emph{Proceedings of the European Conference on Computer Vision
  ({{ECCV}})}.

\bibitem[{Yu et~al.(2022)Yu, Zhang, Ji, and Zhen}]{yuACPAutomaticChannel2022}
Yu, H.; Zhang, W.; Ji, M.; and Zhen, C. 2022.
\newblock {{ACP}}: {{Automatic Channel Pruning Method}} by {{Introducing
  Additional Loss}} for {{Deep Neural Networks}}.
\newblock \emph{Neural Processing Letters}.

\bibitem[{Zhou et~al.(2016)Zhou, Ni, Zhou, Wen, Wu, and Zou}]{zhou2016dorefa}
Zhou, S.; Ni, Z.; Zhou, X.; Wen, H.; Wu, Y.; and Zou, Y. 2016.
\newblock DoReFa-Net: Training Low Bitwidth Convolutional Neural Networks with
  Low Bitwidth Gradients.
\newblock \emph{CoRR}, abs/1606.06160.

\bibitem[{Zhu et~al.(2017)Zhu, Han, Mao, and Dally}]{zhu2017ttq}
Zhu, C.; Han, S.; Mao, H.; and Dally, W.~J. 2017.
\newblock Trained Ternary Quantization.
\newblock In \emph{5th International Conference on Learning Representations,
  {ICLR} 2017, Toulon, France, April 24-26, 2017, Conference Track
  Proceedings}. OpenReview.net.

\end{thebibliography}
\section*{Appendix}
\subsection{Sequential versus Concurrent Joint Policy Search}
Within this experiment, we aim to compare the results of a joint policy with the result of searching compression policies for both compression methods independently.
To do so, we ran the pruning agent first and used the found policy as a starting point for a search using the quantization agent.
We repeated the experiment in opposite order.
The second run does not change the parameters for the compression method predicted by the first run and predicts parameters for the other compression method only.
For the respective first runs, we used $c_1 = 0.5 \cdot (1-c)$ with $c=0.2$ as a target compression rate.
For all runs of the pruning agent within this experiment, we applied the same channel rounding restriction as for the joint agent.
\begin{figure}[h]
    \begin{subfigure}{0.5\textwidth}
        \centering
        \includegraphics[width=\textwidth]{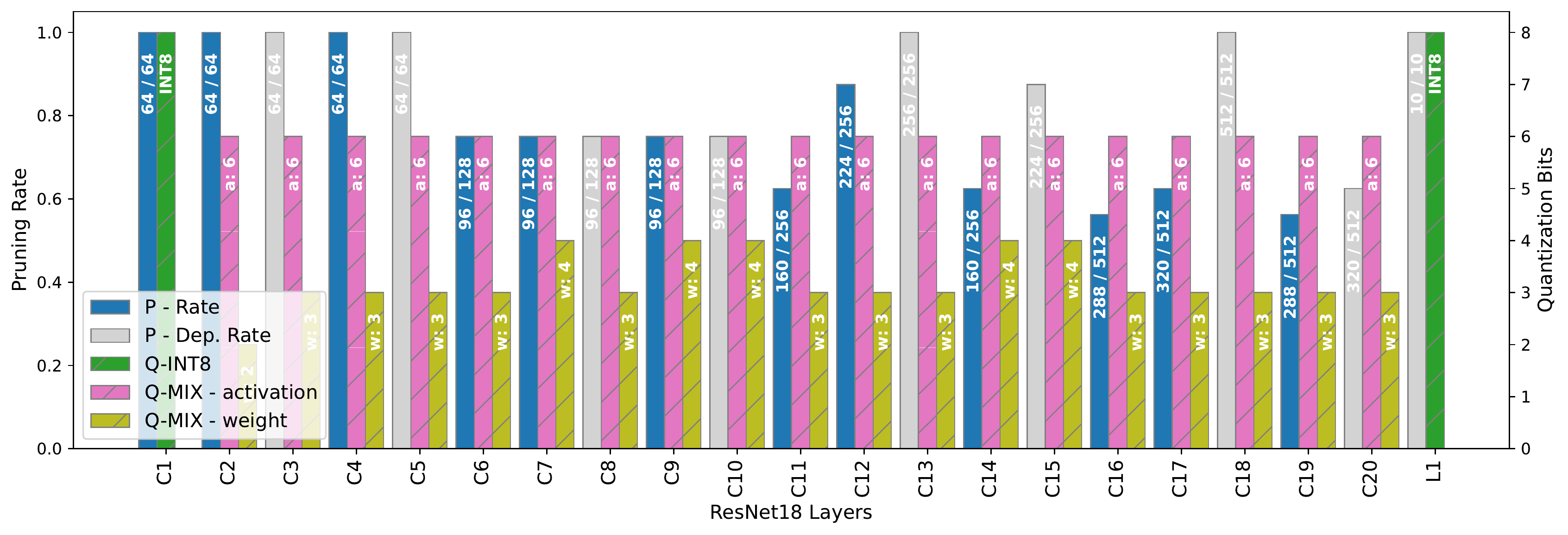}
        \caption{The policy found by a pruning search with $c=0.6$ and subsequent quantization search with $c=0.2$.}
        \label{fig:exp03:pruning_first}
    \end{subfigure}
    \begin{subfigure}{0.5\textwidth}
        \centering
        \includegraphics[width=\textwidth]{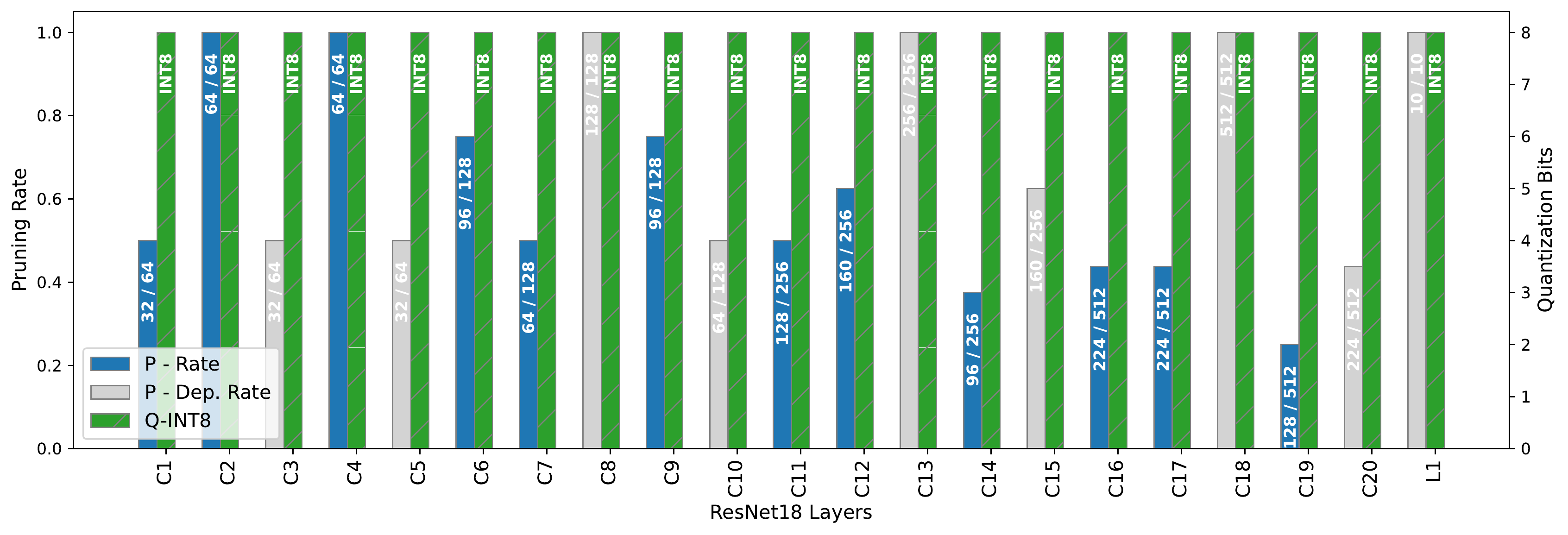}
        \caption{The policy found by a quantization search with $c=0.6$ and subsequent pruning search with $c=0.2$.}
        \label{fig:exp03:quantization_first}
    \end{subfigure}
	\begin{subfigure}{0.5\textwidth}
        \centering
        \includegraphics[width=\textwidth]{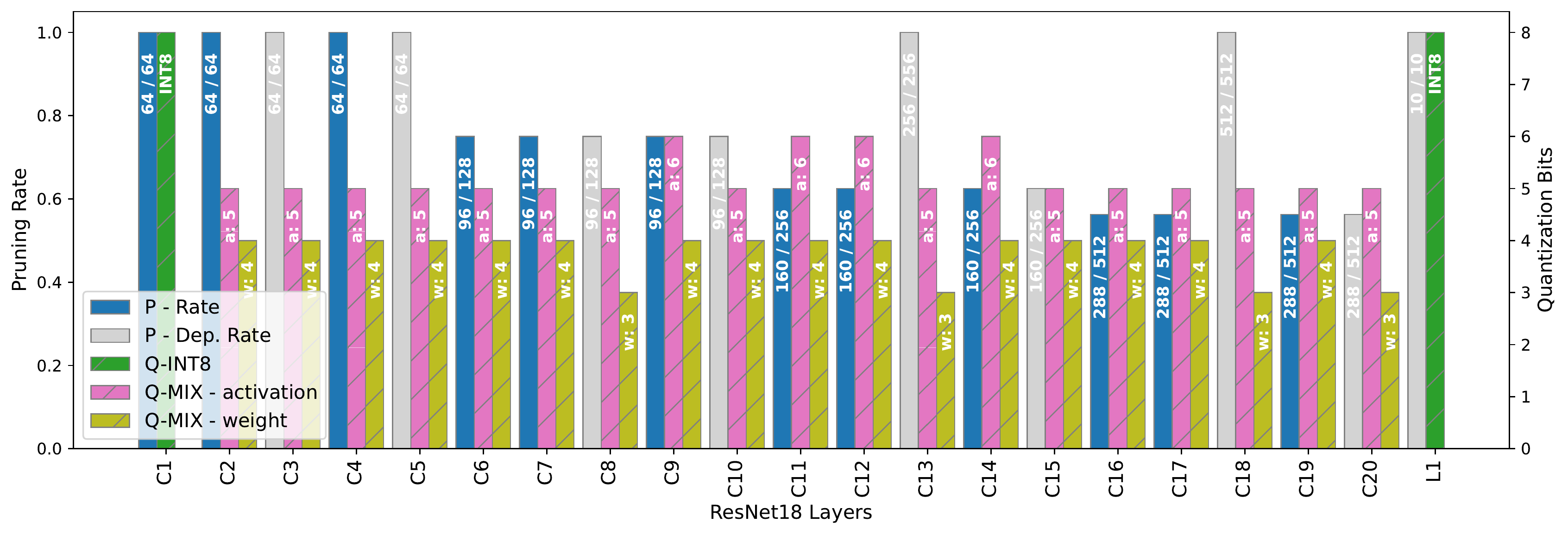}
        \caption{The policy found by a joint search for $c=0.2$.}
        \label{fig:exp03:joint}
    \end{subfigure}
    \caption{Policies found by different search schemes for an effective target compression rate $c=0.2$. The subsequent search strategies use the second compression more aggressively. In contrast the joint search uses both compression method in a balanced way.}\label{fig:exp03:policies}
\end{figure}

\begin{figure*}[h]
    \centering
    \begin{subfigure}{0.3\textwidth}
        \centering
        \includegraphics[width=\textwidth]{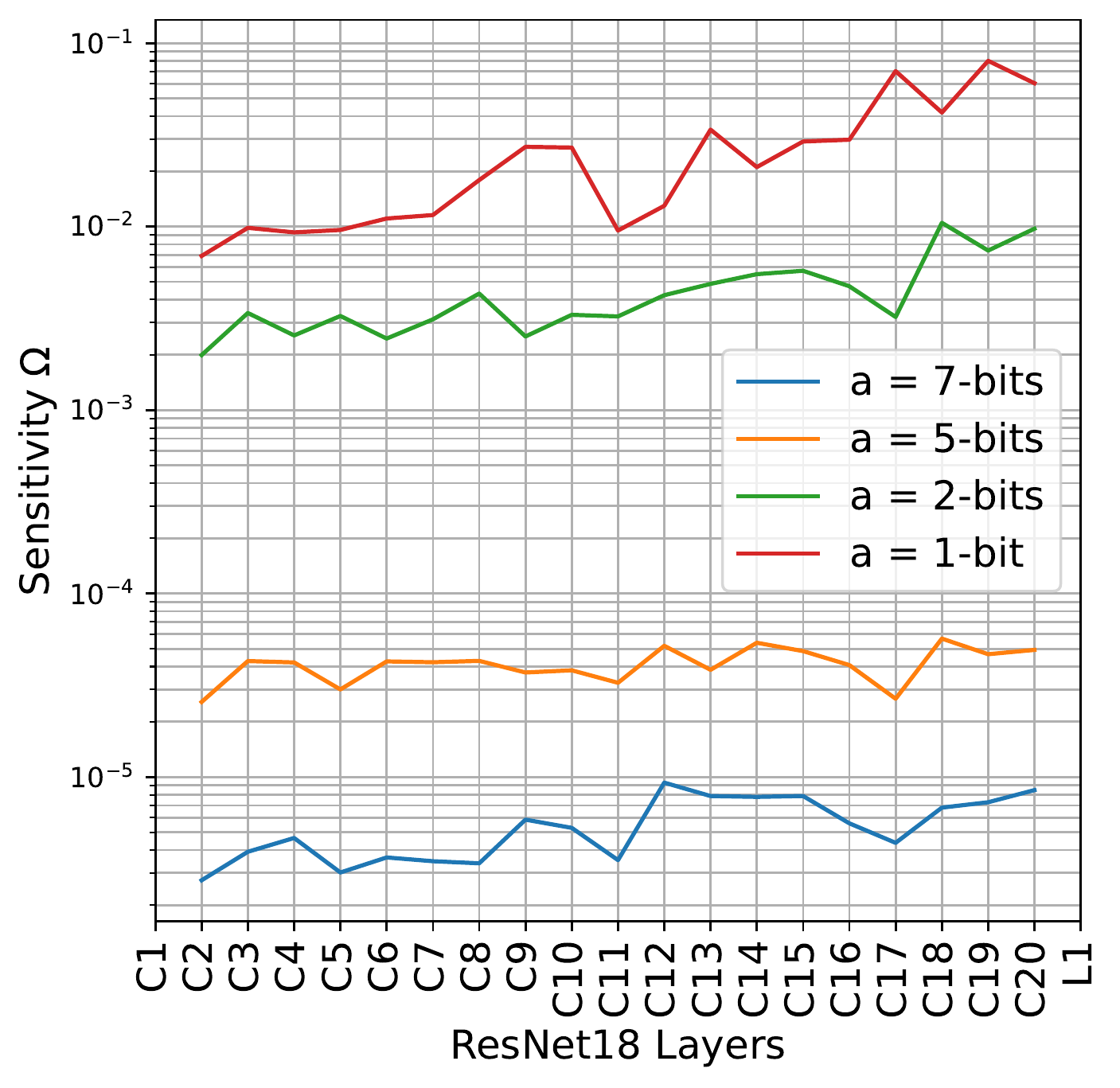}
        \caption{Sensitivity activation}
        \label{fig:eval:sens:activation}
    \end{subfigure}
    \begin{subfigure}{0.3\textwidth}
        \centering
        \includegraphics[width=\textwidth]{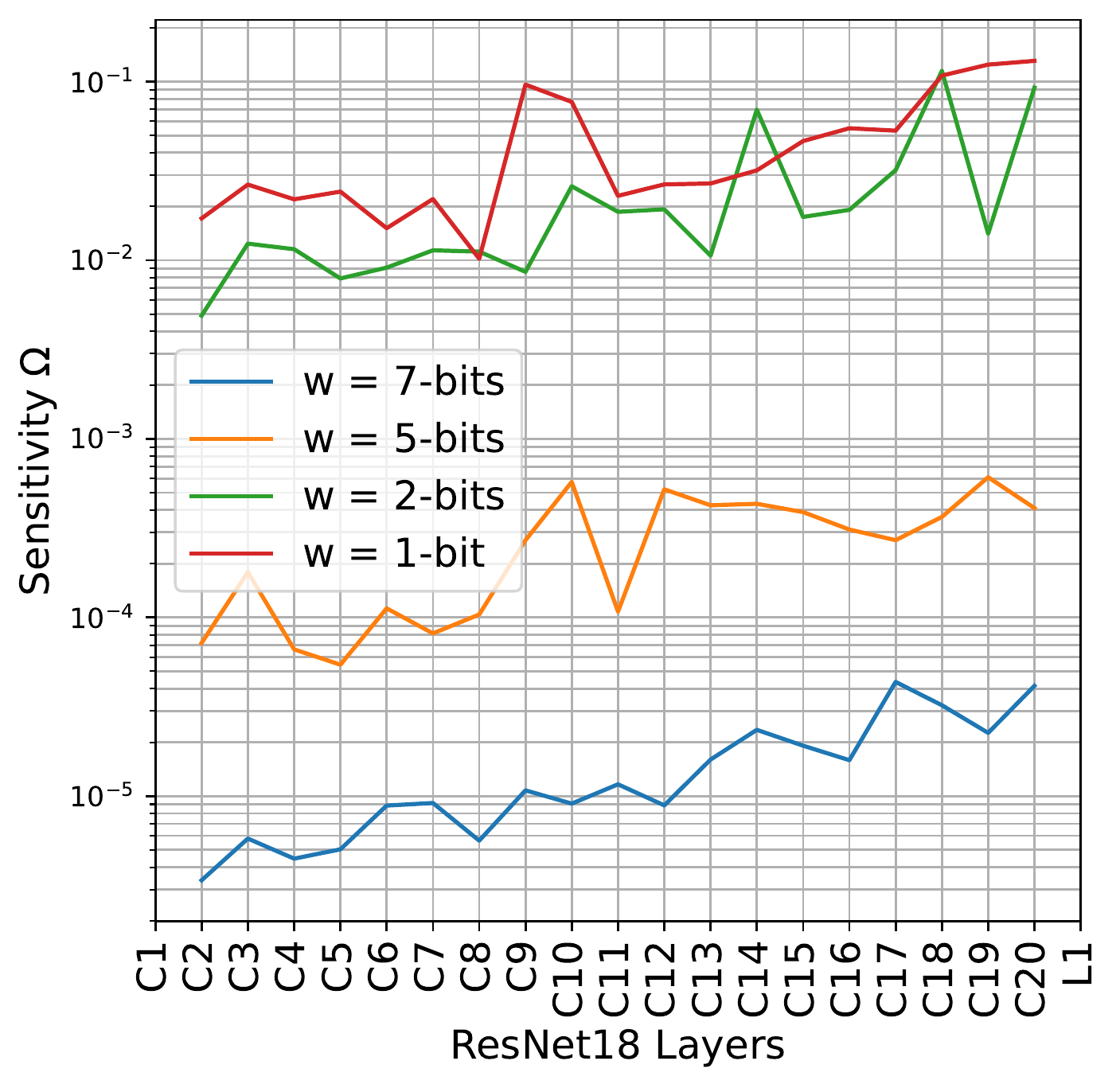}
        \caption{Sensitivity weights} 
        \label{fig:eval:sens:weights}
    \end{subfigure}
    \begin{subfigure}{0.3\textwidth}
        \centering
        \includegraphics[width=\textwidth]{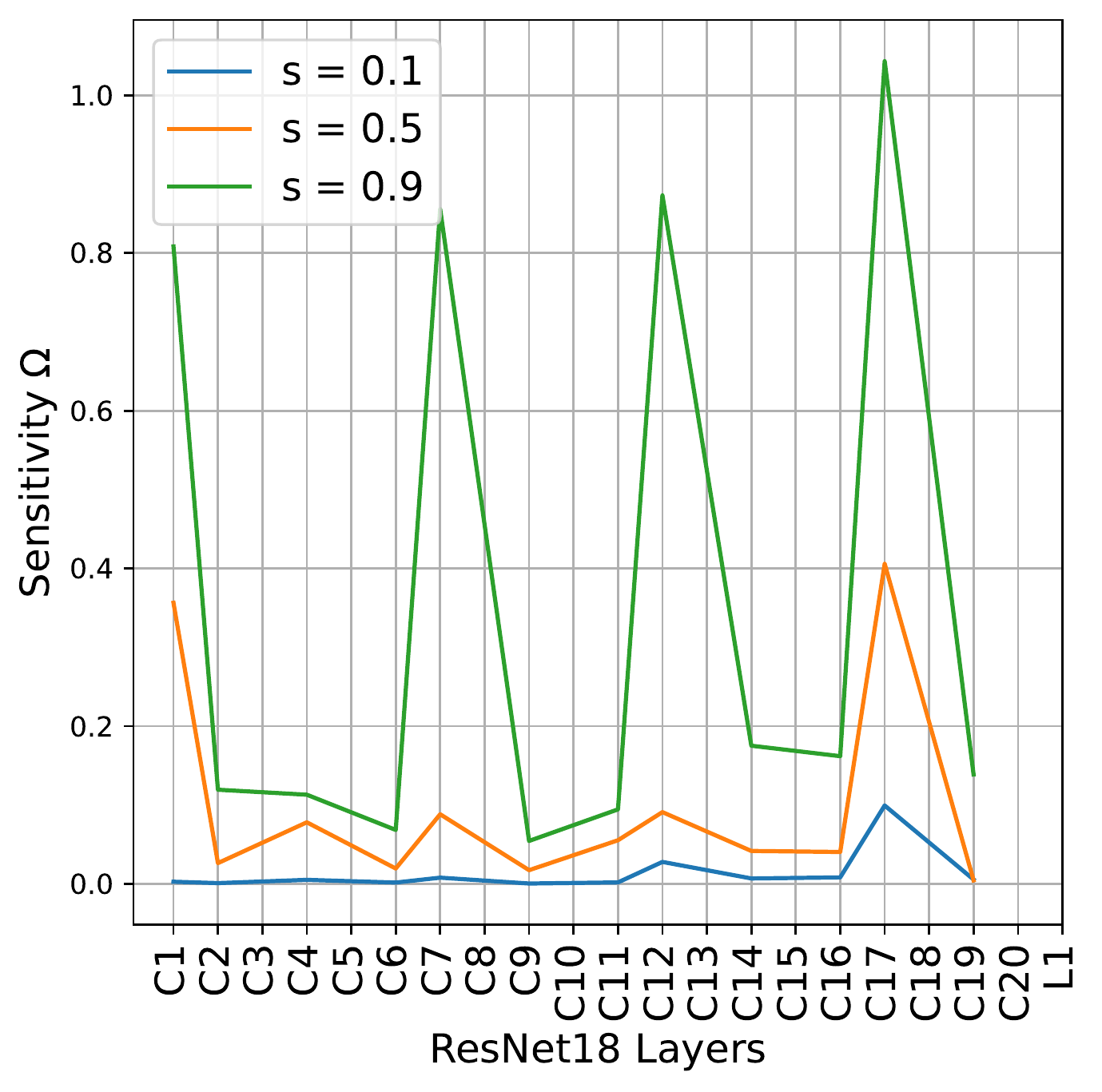}
        \caption{Sensitivity pruning.} 
        \label{fig:eval:sense:sparsity}
    \end{subfigure}
    \caption{Sensitivity over layers:  To activation and weight quantization and to column pruning.}\label{fig:eval:sens:aw}
\end{figure*}

Figure \ref{fig:exp03:policies} shows the joint policies found by all three search schemes.
The pruning actions included in the policies found by the pruning first and the joint search scheme are quite similar.
For quantization, the split search scheme used less restrictive activation quantization but therefore reduced the weight precision more aggressively.
Contrarily, the quantization-first search scheme found a different policy.
The quantization agent predicted the least aggressive \textbf{INT8} quantization option for all layers.
The subsequent pruning run applied much more aggressive pruning actions to meet the compression target, even involving a pruning of the first layer, which was not touched by the other two search schemes.

With that, it seems that the search schemes using two separate runs lead to a more aggressive usage of the compression method corresponding to the second run.
Although both runs have the same latency budget, the complexity of the task might not scale linearly with the target compression rate.
The joint agent could balance between both methods much better and therefore achieves the same result with less restrictive compression.

\subsection{Influence of the Sensitivity Metric}
One central enhancement of our algorithm compared to AMC and HAQ is we run a sensitivity analysis and include the results in the agent states.
The sensitivity metric should measure the impact of each compression method per layer.
During the analysis, we compare the output of an uncompressed and compressed model by measuring the Kullback-Leibler divergence.

Figure \ref{fig:eval:sens:aw} shows the sensitivity per layer for mixed precision quantization with varied activation or weight bit width.
Both parameters are tested independently, the respective counterpart is set to the maximum bit width allowed.
We selected a subset of bit widths for the plotting---for searches all possible bit widths are tested.
Clear differences in sensitivity are observable for each layer.
Overall each bit width results in another level of sensitivity, thus lower bit widths result in higher sensitivity.
For both, activation and weights, a clear trend towards higher sensitivity for later layers is visible.
With that quantizing later layers stronger should lead to a greater loss in accuracy.

Figure \ref{fig:eval:sense:sparsity} shows the sensitivity results of different sparsities for all layers.
The sparsity is discretized to a concrete channel count.
For the sensitivity analysis we sample 10 test points uniformly across the sparsity range.
For pruning an even clearer pattern of more and less sensitive layers is observable. 

To validate the importance of the sensitivity analysis for our algorithm we conducted an ablation study.
We ran a search using the joint agent with a target compression rate $c=0.2$ with disabled sensitivity analysis.
For all sensitivity-based features within the agent state a constant value was set.

\begin{table}[h]\renewcommand{\arraystretch}{1.2} 
	\caption{Overview of quantitative results joint searches with enabled and disabled sensitivity analysis for $c=0.2$.}
	\centering
	\small
	\begin{tabular}{l|r|r|r|r}
	\bfseries   & \multicolumn{1}{c|}{\bfseries MACs} & \multicolumn{1}{c|}{\bfseries BOPs} &  \multicolumn{1}{c|}{\bfseries Rel. Lat.} & \bfseries Accuracy \\ \hline
	Reference 	& $4.75 \cdot 10^{10}$          & $4.86 \cdot 10^{13}$ 		& $100.0$\,\% 	& $93.04$\,\% \\
	Disabled 			& $1.68 \cdot 10^{10}$ 	& $8.10 \cdot 10^{11}$ 		& $20.0$\,\% 	& $92.66$\,\% \\
	Enabled 		    & $2.82 \cdot 10^{10}$  & $6.75 \cdot 10^{11}$ 		& $19.4$\,\% 	& $92.77$\,\% \\
	\end{tabular}
	\label{tab:exp:sens:overview} 
\end{table}

Table~\ref{tab:exp:sens:overview} shows an overview of the search results compared to a search with enabled sensitivity analysis.
The search with enabled sensitivity found a policy resulting in marginal higher accuracy.
Despite this, it is observable that the disabled search used pruning more aggressively---the number of MACs is substantially lower.
It seems that quantization was used less instead.
Still, the agent was able to meet the given latency constraint.
\begin{figure}
    \centering 
    \begin{subfigure}{0.5\textwidth}
        \centering
        \includegraphics[width=\textwidth]{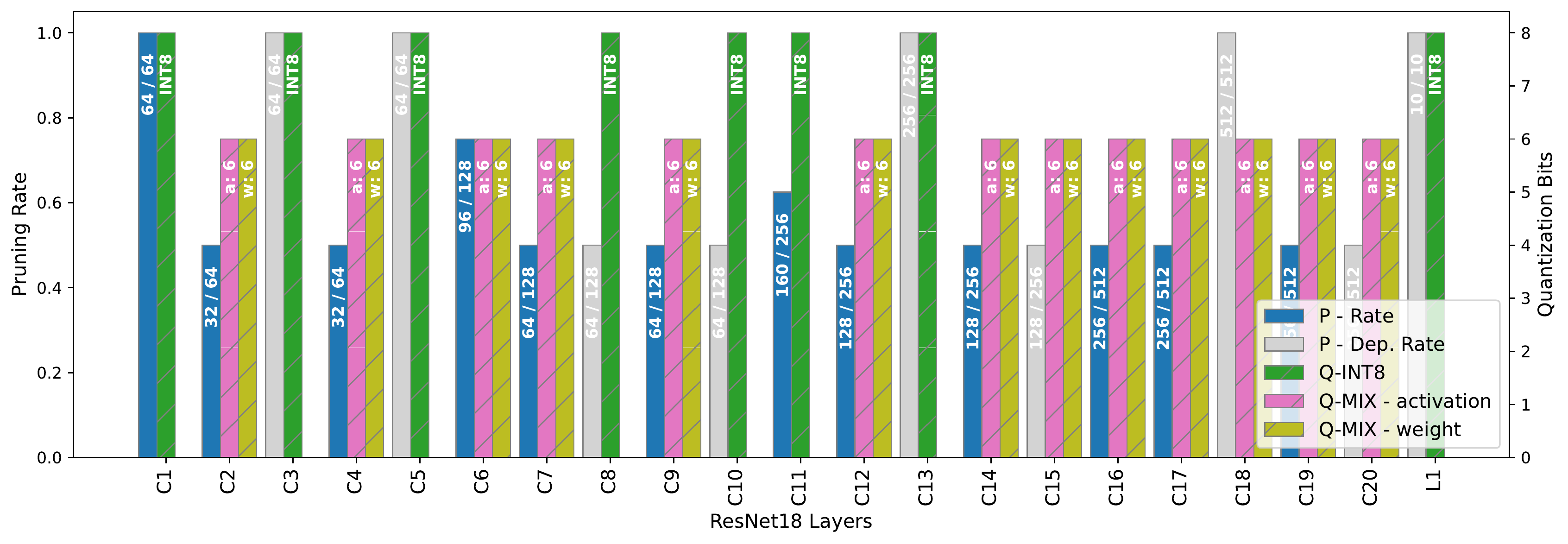}\label{fig:exp:sens:no_sens}
        \caption{Without Sensitivity Feature}
    \end{subfigure}
    \begin{subfigure}{0.5\textwidth}
        \centering
        \includegraphics[width=\textwidth]{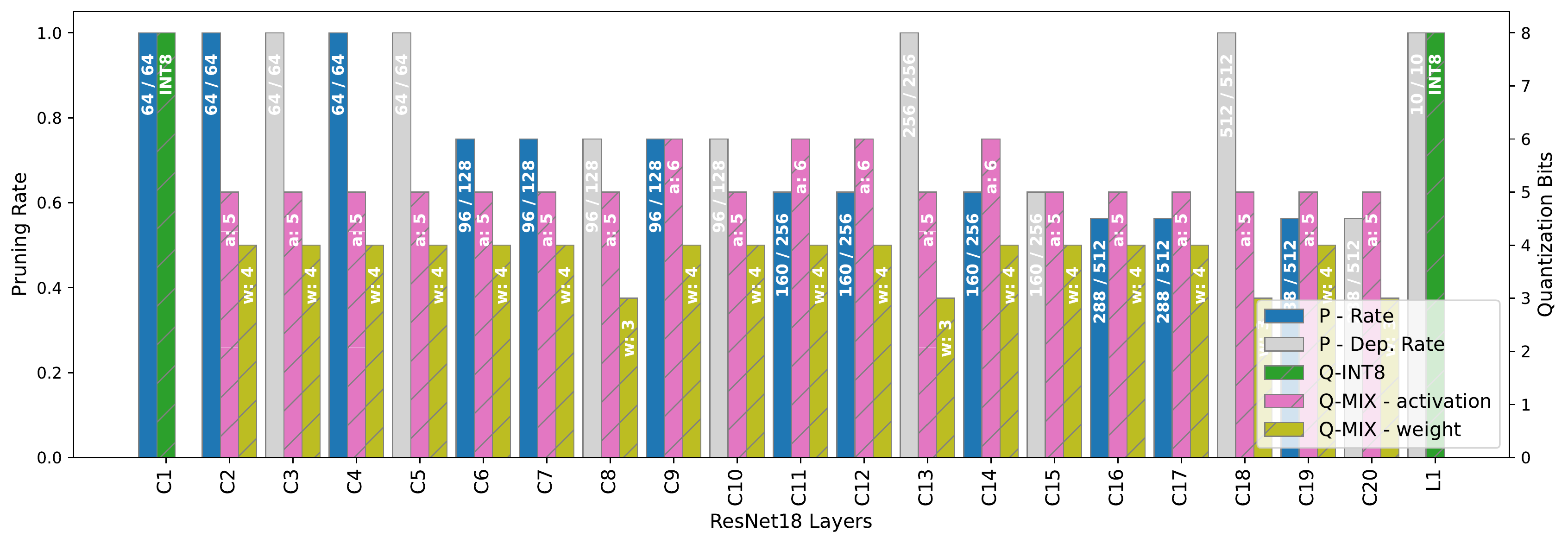}\label{fig:exp:sens:sens}
        \caption{Using Sensitivity Feature}
    \end{subfigure}
    \caption{Joint policies found for target compression rate $c=0.2$ with disabled $(a)$ and enabled $(b)$ sensitivity analysis.}
    \label{fig:sens}
\end{figure}

Figure~\ref{fig:sens} compares the policies found by the searches, here it is clearly visible that: With a disabled sensitivity analysis pruning is used more aggressively.
For quantization, even the integer 8-bit option was used often. 
Without sensitivity analysis the mixed quantization option was used with a precision of 6 bits for activation and weights only.
The agent does not vary the activation or weight bit width across the mixed operations.
An inspection of the continuous actions predicted by the agent showed that the values have low variance across layers.
Instead, the agent predicted similar values for all layers, for weight and activation the value is around the mixed precision threshold.
This indicates that a policy search with sensitivity analysis is beneficial to compress the most resilient layers most and thus exploit heterogeneity in model architectures.
\end{document}